\documentclass[10pt,twocolumn,letterpaper]{article}

 \usepackage[pagenumbers]{cvpr} %

\usepackage[dvipsnames]{xcolor}

\definecolor{cvprblue}{rgb}{0.21,0.49,0.74}
\usepackage[pagebackref,breaklinks,colorlinks,citecolor=cvprblue]{hyperref}

\usepackage{graphicx}
\usepackage{booktabs}
\usepackage{microtype}
\usepackage{algorithm}
\usepackage{algorithmicx}
\usepackage[noend]{algpseudocode}
\usepackage{scontents}
\usepackage{multicol}
\usepackage{multirow}
\usepackage{float}
\usepackage{wrapfig}
\usepackage[noend]{algpseudocode}
\usepackage{soul}
\usepackage{changepage,threeparttable} %
\usepackage{array}
\usepackage{adjustbox}

\def\paperID{*****} %
\def\confName{3DV\xspace}
\def\confYear{2024\xspace}
\newcommand{\methodname}{ObjectCarver}
\newcommand{\bh}[1]{\textcolor{red}{#1}}
\title{\methodname: Semi-automatic segmentation, reconstruction and\\ separation of 3D objects}

\author{Gemmechu Hassena, Jonathan Moon, Ryan Fujii, Andrew Yuen,\\
Noah Snavely, Steve Marschner, Bharath Hariharan \\ 
Cornell University
}

\begin{document}

\makeatletter
\let\@oldmaketitle\@maketitle%
\renewcommand{\@maketitle}{

\@oldmaketitle%
    \captionsetup{type=figure}
    \centering
    \vspace{-0.5cm}
  \includegraphics[width=0.9\linewidth]
    {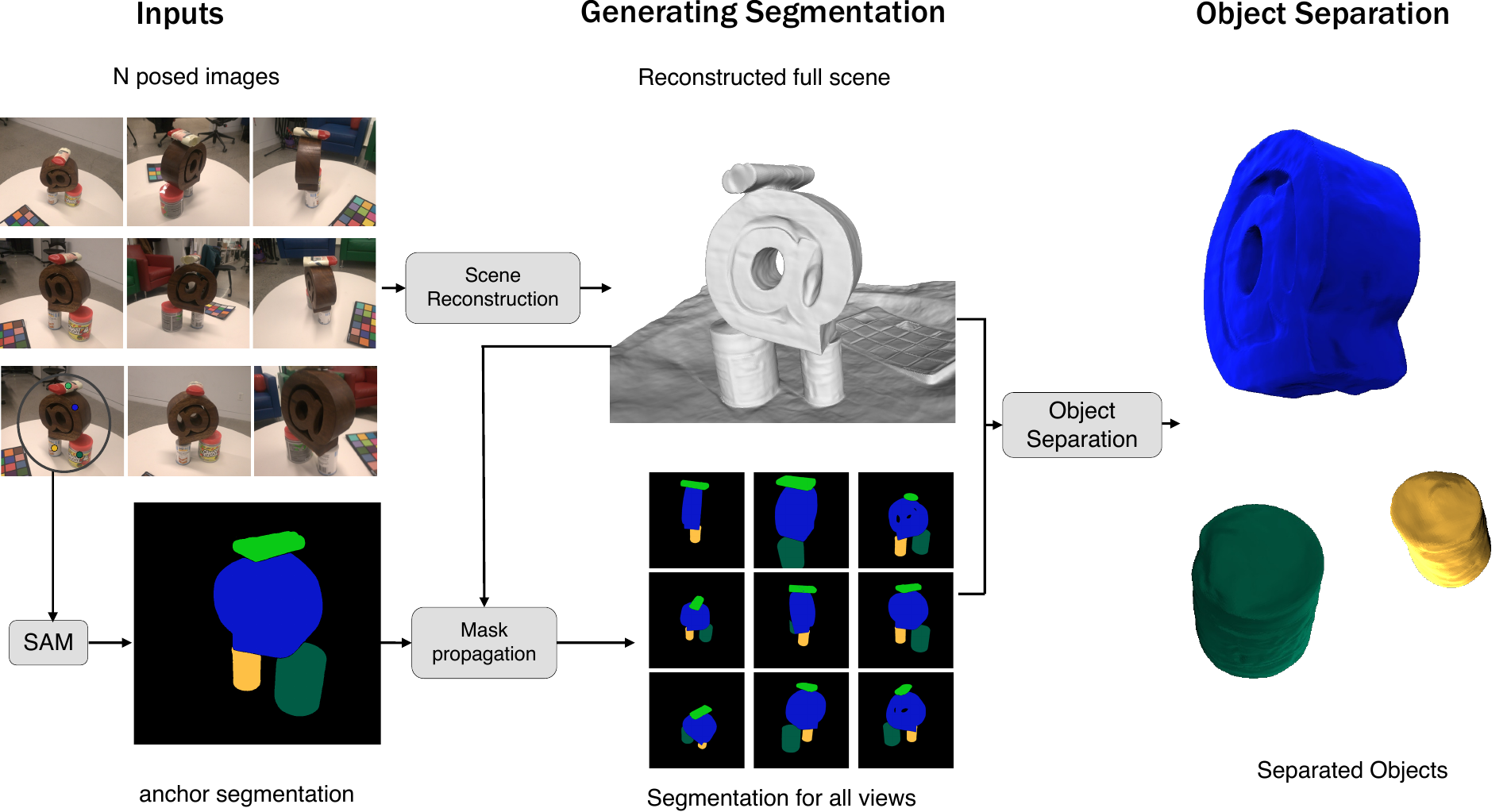} 
    \caption{Our approach takes multiple views of a scene as input, along with a few point clicks on one of the views, which can be converted into segmentation masks (left). It then: (a) segments all the other images, and (b) reconstructs each segmented object, completing the occluded regions if any. }
    \label{fig:intro}
    \bigskip
    }
    
\makeatother

\maketitle

\begin{abstract}
Implicit neural fields have made remarkable progress in reconstructing 3D surfaces from multiple images; however, they encounter challenges when it comes to separating individual objects within a scene. Previous work has attempted to tackle this problem by introducing a framework to train separate signed distance fields (SDFs) simultaneously for each of N objects and using a regularization term to prevent objects from overlapping. 
However, all of these methods require segmentation masks to be provided, which are not always readily available. 
We introduce our method, {\methodname}, to tackle the problem of object separation from just click input in a single view. 
Given posed multi-view images and a set of user-input clicks to prompt segmentation of the individual objects, our method decomposes the scene into separate objects and reconstructs a high-quality 3D surface for each one. 
We introduce a loss function that prevents floaters and avoids inappropriate carving-out due to occlusion.  
In addition, we introduce a novel scene initialization method that significantly speeds up the process while preserving geometric details compared to previous approaches.
Despite requiring neither ground truth masks nor monocular cues, our method outperforms baselines both qualitatively and quantitatively. In addition, we introduce a new benchmark dataset for evaluation. 

\end{abstract}
    
\section{Introduction}
\label{sec:intro}
With recent advances in neural implicit scene representations, we can now reconstruct 3D scenes with complete, high-quality surfaces (represented as signed distance functions or SDFs) from a set of images taken by cameras with known poses \cite{neus,volsdf}.
Although these techniques compute high-quality surfaces, they are limited to representing the entire scene as a single surface. 
This representation is fine for applications such as walkthroughs where the scene remains fixed,
but for many applications it is 
desirable to extract and manipulate individual objects, including applications in robotics and virtual reality where simulating such scene manipulations is crucial.
In this paper, we tackle this problem of 3D scene decomposition: 
given multiple views of a 3D scene, can we produce a reconstruction where the individual objects are separated out?

\begin{figure}[H]
  \centering
   \includegraphics[width=1\linewidth]{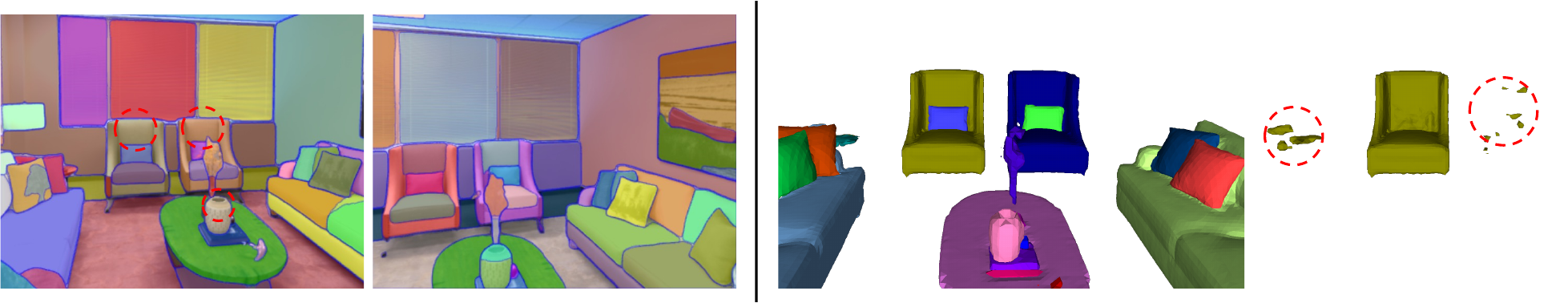} %
   \caption{\textbf{Failure cases of SOTA.} 
   Using SAM independently on each image precludes corresponding objects between views (Left). Even if one were to solve this correspondence problem, slight errors in SAM output mean that the same object may be segmented differently in the different views (e.g., the top of the vase is included in the vase segment in the left image but not the right). Even with good segmentations, prior work such as ObjectSDF++ \cite{objsdf++} introduces floating artifacts, especially those hidden behind other objects (Right).}
   \label{fig:autosam}
   \label{fig:failure}
\end{figure}

Some previous works \cite{objcompnerf,objsdf,objsdf++,li2023rico} have  addressed the problem of reconstructing many separate objects.
However, two key challenges remain.
First, these techniques require 
segmentation masks of each object in each view
as part of the input.
Unfortunately, the cost of the manual work involved in producing such segmentations scales with the number of input views and the number of objects, making the process cumbersome.
Automated solutions like the Segment Anything Model (SAM) \cite{sam} often over-segment and result in inconsistent segmentation across multiview images (Fig.~\ref{fig:autosam}, left). Recent works, such as SA3D \cite{sa3d}, that attempt this problem use volume density, but our method uses SDF, where we know exactly where the surface lies.

Second, prior work fails in the presence of occlusion (Fig.~\ref{fig:autosam}, right).
Parts of the scene that are occluded from all views provide no supervision for existing techniques, giving the model free rein to introduce floating components in the occluded regions.
These floating artifacts can be large and numerous and as such result in extremely inaccurate object reconstructions.

We introduce \methodname{} to address these limitations.  
\methodname{} takes as input a collection of posed images and point clicks of each object for segmentation in just \emph{one} of the views. This first segmentation can be performed with tools like SAM~\cite{kirillov2023segment}.
\methodname{} then outputs object segmentations for all input views and a high-fidelity 3D surface for each object (Figure~\ref{fig:intro}). 
This 3D surface includes not just the parts of the object that are visible but also makes reasonable completions in completely occluded regions where no image evidence is available.
Crucially, \methodname{} removes almost all floating artifacts that plague prior work.
Finally, \methodname{} achieves this reconstruction with a fairly small computational overhead beyond the computational cost of full scene reconstruction.

\methodname{} works in three phases. First, we reconstruct the entire 3D scene as a single SDF using existing methods \cite{neus}. Then, from one segmentation mask (computed from the user's input click using SAM \cite{sam}), we use the reconstructed 3D surface together with SAM~\cite{kirillov2023segment} to propagate segmentation labels to the other input images, resulting in accurate and multi-view consistent masks for each object. Finally, we jointly train per-object SDF surfaces, starting from the full-scene SDF.
We introduce a novel loss function to produce a set of consistent and compact 3D surfaces.

Finally, we find that existing benchmarks for this task are limited, with incomplete ground-truth object meshes and metrics that do not correctly penalize floaters.
Therefore, we introduce a new dataset of both synthetic and real-world scenes consisting of multiple objects and equipped with a ground-truth mesh for each object.
We also introduce updated metrics that correctly penalize all error modes.
We compare our method with prior methods both qualitatively and quantitatively in this benchmark and demonstrate that our method outperforms the previous methods for this problem.
In sum, our contributions are:

\begin{enumerate}
    \item A new {\bf automatic segmentation} approach that leverages the 3D scene structure to generate object segmentations for all the input images from just a few points the user clicks in one view.
    \item A new {\bf object compactness loss} that removes floaters in occluded regions and produces substantially more accurate reconstruction; and
    \item A change of {\bf initialization} for the object models that improves surface quality and considerably speeds up convergence.
    \item \textbf{Synthetic and real-world datasets} of multi-object compositional scenes and their individual geometries.

\end{enumerate}

\section{Related Work}
\label{sec:related}

\textbf{Neural field representations for geometry.} Neural representations for surface geometry began with methods that trained using 3D supervision \cite{OccupancyNet, DeepSDF}, but soon began to focus on using more readily available multi-viewpoint images as supervision \cite{IDR, DVR}.  Neural Radiance Fields \cite{NeRF} introduced a framework to use volumetric rendering to train radiance fields, leading to follow-on work improving training and rendering speed \cite{KiloNeRF, PlenOctrees, NGLOD, iNGP}, handling complex, unbounded, and dynamic scenes \cite{nerf++, mipnerf360, nerfies, hypernerf, nsff, dynibar}, and improving representation quality \cite{mipnerf, zipnerf}.

To obtain more explicit geometric representations than NeRFs provide, some recent advances have optimized neural signed distance functions (SDFs) by using them to define smooth volume densities that are rendered in the NeRF framework, which helps guide the training process stably to accurate and detailed surfaces.  VolSDF \cite{volsdf} and NeuS \cite{neus} both achieve good surface reconstructions in this way; building on these methods, MonoSDF \cite{monosdf} incorporates monocular cues and PermutoSDF \cite{rosu2023permutosdf} achieves detailed reconstructions of small-scale features.

\medskip
\noindent \textbf{Decomposing 3D scenes into objects.}
The methods above focus on reconstructing geometry or radiance fields, but do not attempt to further understand scenes as compositions of objects. A number of methods for disentangling separate objects have been proposed. Some of these methods learn from observing scenes without further supervision. 
Niemeyer and Geiger proposed GIRAFFE \cite{giraffe}, which utilizes latent codes to generate object-centric Neural Radiance Fields (NeRFs) and conceptualize scenes as compositional generative neural feature fields. uORF learns unsupervised object composition models that can be used to factor new scenes at inference time~\cite{yu2022unsupervised}.  
DiscoScene \cite{xu2022discoscene} uses weak supervision in the form of \emph{layout prior} for object-compositional generation but fails to generalize to unknown objects.
In contrast to the high-level object decompositions of the above work, Differentiable Blocks World \cite{dbw} trains a mid-level scene representation from multiple images. Rather than achieving the highest geometric quality, that method aims to decompose the scene into mid-level 3D textured primitives.

Other work uses joint language-visual embeddings like CLIP to identify objects in 3D scenes. Sosuke \etal use CLIP and DINO to learn neural feature fields, supporting editing and selection mechnisms~\cite{kobayashi2022decomposing}. LERF~\cite{lerf} learns a language field by volumetrically rendering proto-CLIP features along the ray which is supervised with multi-scale CLIP features on the training images, allowing radiance fields to be decomposed into semantically distinct areas.

In contrast to CLIP, our method relies on a pre-trained 2D image segmentation network. 
Other work in this vein includes ObjectNeRF, which separates scenes into disjoint radiance fields for each object based on rough 2D instance masks~\cite{objcompnerf}. 
More recently, the emergence of the Segment Anything Model (SAM) marked a significant step towards segmenting 2D images~\cite{kirillov2023segment}. Extending this model to 3D object segmentation, Segment Anything 3D (SA3D) \cite{sa3d} uses mask inverse rendering and cross-view self-prompting to construct 3D masks, demonstrating adaptability to various scenes and efficiency in achieving 3D segmentation. However, unlike our method, SA3D segments a fixed 3D representation and does not attempt to \emph{separate} objects from one another, i.e., to modify their geometry to, e.g., fill in holes at interfaces where they are in contact.

Another key difference with the above work is that we seek not to produce segmented NeRFs, but instead segmented, separated, and high-quality \emph{surfaces} in the form of SDFs that can be converted into convenient graphics representations like meshes. In that sense, our work is similar to ObjectSDF \cite{objsdf}, which uses per-image input instance masks to product an SDF for each object.
However, this method can encounter issues with object and scene reconstruction accuracy, slow convergence, and training speed. Its successor ObjectSDF++ \cite{objsdf++} introduces an occlusion-aware object opacity rendering strategy and an overlap regularization term to better separate the surfaces between neighboring objects. However, it still requires per-image, per-object input masks, in contrast to our method. RICO \cite{li2023rico} leverages geometrically motivated regularizations to smooth unobserved regions in indoor compositional scenes, whereas our method goes farther to separate and reconstruct complete objects.
Our method is in the spirit of other semi-supervised methods like that of Ren \etal~\cite{ren2022nvos}, but scales well to complex scenes with many objects.

\section{Methodology}
\label{sec:method}

\begin{figure*}[t]
  \centering
   \includegraphics[width=0.8\textwidth]{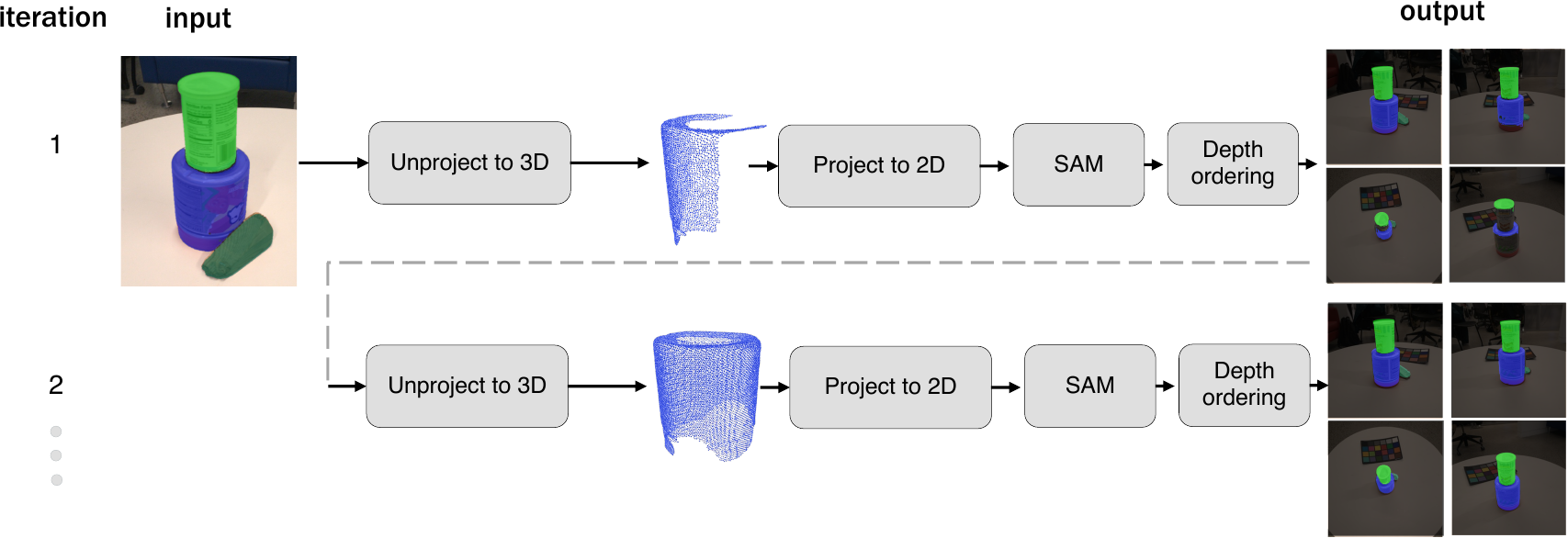}

   \caption{\textbf{Mask Propagation pipeline:} in the first iteration, a user clicks a point on each object to generate a per-object anchor mask, which are then unprojected into 3D (here, we only show unprojected 3D points for the bottom can). 
   These 3D points are subsequently projected back into each image view, while checking for occlusions. The projected points serve as seeds for SAM \cite{sam} to generate masks for each object (bottom and top cans, door stop). To combine these individual segmentation masks into a single image, we use a depth ordering technique. In the next iterations, all views are used as anchor masks, allowing the pipeline to cover previously unseen regions.}
   \label{fig:mask_prop}
\end{figure*}

We assume that we are given a set of $N$ posed images $\mathcal{I} = \{I_1, \ldots, I_N\}$ of a scene.
We are interested in not just reconstructing the scene, but segmenting, reconstructing and \emph{separating} each of $K$ different objects in the scene.
By separation, we mean to produce an SDF representation of each of the $K$ objects so that they can be manipulated at will.
We aim to do so as accurately, as efficiently, and with as little manual annotation as possible.

Our proposed approach operates in three stages:
\begin{enumerate}
\item Reconstruct the full scene as a single SDF.
\item Generate segmentation masks for each of the $K$ objects in all images by propagating segmentation mask from one of the views.
\item Optimize $K$ separate SDFs using a novel loss to handle occlusion and contacts between objects for accurate reconstruction.
\end{enumerate}
We next describe each step below.

\subsection{Scene Reconstruction} 
We first train a full scene reconstruction. Any SDF-based technique can be used; however, here we use NeuS~\cite{neus} which converts the SDF into a density term to allow for optimization through volumetric rendering.
Concretely, for every pixel $\mathbf{p}$, discrete samples are taken along the corresponding ray $\{\mathbf{p}_i = \mathbf{o} + t_i \mathbf{v} | i = 1, \ldots n, t_i < t_{i+1}\}$ where $\mathbf{o}$ is the camera center and $\mathbf{v}$ is the viewing direction corresponding to the pixel.
Then NeuS calculates densities $\alpha_i$ and an accumulated transmittance $T_i = \prod_{j=1}^{i-1}(1-\alpha_i)$.
The density is shown to be related to the SDF as:
\begin{align}
\alpha_i = \max \left(\frac{\Phi_s\left(f(\mathbf{p}_i)\right) - \Phi_s\left(f(\mathbf{p}_{i+1})\right)}{\Phi_s\left(f(\mathbf{p}_i)\right)}, 0 \right)
\end{align}
where $\Phi_s$ is the sigmoid function and $f$ is the SDF.
(Please refer to Wang et.\ al \cite{neus} for details.)
Given these densities $\alpha_i$ and the corresponding accumulated transmittance $T_i =  \prod_{j=1}^{i-1} (1-\alpha_j)$, the rendered color at this pixel is computed as:
\begin{align}
C(\mathbf{o}, \mathbf{p}) = \sum_i T_i \alpha_i c(\mathbf{p}_i, \mathbf{v})\label{render}
\end{align}
where $c(\mathbf{p}_i, \mathbf{v})$ is the color at the point $\mathbf{p}_i$ seen from the viewing direction $\mathbf{v}$.

The SDF is optimized to minimize rendering and eikonal losses: 
\begin{equation}
    L = L_{\text{color}} + \lambda L_{\text{eik}} 
\end{equation}
\begin{equation}
    L_{\text{color}} = \frac{1}{m} \sum_{j} \|\hat{C}_j - C_j\|
\end{equation}
\begin{equation}
    L_{\text{eik}} = \frac{1}{nm} \sum_{j,i} (\| \nabla f(\textbf{p}_{j,i}) \|_2 - 1)^2
\end{equation}

Here $j$ indexes over pixels and $i$ indexes over points sampled along a ray.
$\hat{C}_j$ is the predicted color, $C_j$ is the observed color, $m$ is the number of pixels, and $n$ is the number of samples per ray, and $\textbf{p}_{j,i}$ is the sampled point along pixel $j$ at index $i$.

\subsection{Generating Segmentations}
\label{sec:segment}

Our next step is to segment each of the $K$ objects in each of the $N$ images. Given a few clicked points for one of the images, we use SAM~\cite{kirillov2023segment} to generate the segmentation; we call this our anchor mask. Then we use the reconstructed 3D scene to backproject the mask into 3D, resulting in labeled 3D points for each object. Using these labeled 3D points we propagate the segmentation to all views. Finally, we iterate through this process again, using the newly obtained segmentation as the anchor mask (for two iterations, in our implementation). Below we describe each step in detail.

\begin{figure}
\centering

\includegraphics[width=1  \linewidth]{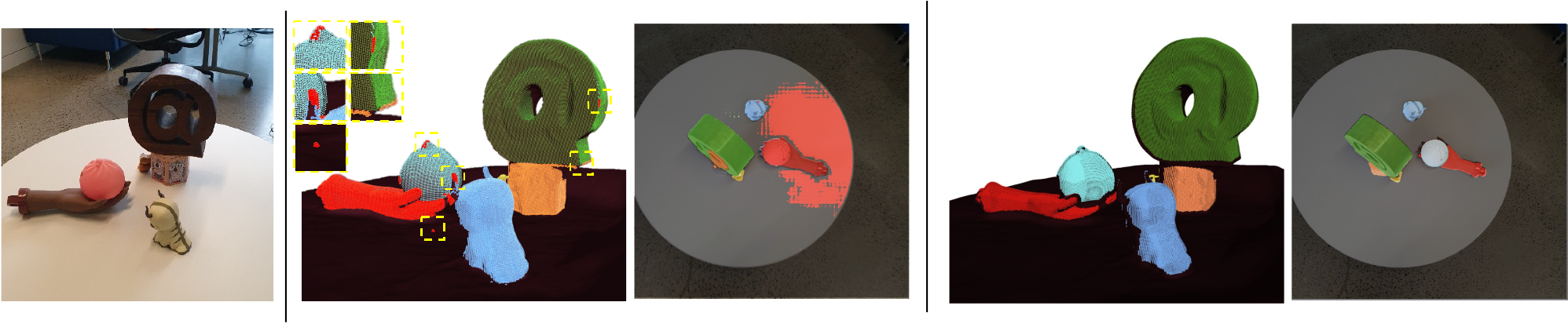}
\caption{\textbf{Projection to 3D.} Left: Example image. Middle: points projected without mask edge erosion and outlier removal, resulting in noisy segmentation outputs. Right: by using mask erosion and outlier removal we obtain clean 3D points and subsequently obtain a correct segmentation output. 
}

\label{fig:outlier}
\end{figure}

\textbf{3D point labeling: }
After generating the anchor mask, we project it into 3D by tracing rays from each pixel through the object mask to determine surface intersections (Figure \ref{fig:mask_prop}). However, segmentations can often be imprecise near object boundaries, causing the mask to leak onto other surfaces (Figure \ref{fig:outlier}). To address this, we first erode the mask to 
remove any segmentation errors where the mask overshoots the true boundary. Second, after back-projecting the points to 3D we remove all points from an object mask whose depths are outliers, i.e., more than $\theta$ standard deviations from the mean depth of the object. We found 2.5 to be a good threshold in our experiment.

Finally, we ensure that each 3D point has a unique label 
by discarding points with more than one label.

\textbf{Propagating to a new view: } 
To segment an object in a new view, we project the labeled 3D points into that view (as long as they are unoccluded) to obtain labeled 2D image points (``seeds''). In principle, these 2D points can be used to  prompt SAM. However, in practice, SAM tends to oversegment when prompted with numerous seeds.
To avoid this, we employ a coreset selection algorithm (algorithm in the supplementary) to reduce the seed points while preserving the object's shape. 

Finally, to reconcile multiple overlapping segmentations, we perform a partial ordering of the different objects based on depth.
We compare the depths of the seed points of each mask in the overlapping areas, and assign the overlapping area to the object that is closer.
For example, in Figure~\ref{fig:mask_prop}, this depth ordering allows us to correctly place the green can pixels in front of the blue can when viewed from the top.
Note that this approach assumes that a depth ordering exists for every pair of objects, which can be false if objects are intertwined, but it works in the vast majority of cases. Please refer to the supplementary for detail.
\subsection{Object Separation}
\label{sec:object}
Given the images and their segmentation masks, our goal is to now produce $K$ separated SDFs, one per object. 
We can train the $K$ SDFs by updating the color loss so that each SDF is only responsible for producing the colors of the corresponding object:  
\begin{equation}
    L_{\text{color}} = \frac{1}{m} \sum_{k}\sum_{j} M_{k} (j) \|\hat{C}_j - C_j\|
\end{equation}

However, this is not enough to separate out the object
 because the segmentation mask only covers the visible part of the object. When a pixel is not part of the mask, it is ambiguous whether this is because the pixel is outside the extent of the object or because it is occluded.
Furthermore, there are still parts of the scene that are not visible from any of the images. These ambiguous regions include both interfaces between objects and occluded parts of the scene (e.g., below the table in table-top scenes).
It is unclear how these occluded regions of the SDF must be optimized.

In what follows, we first discuss the simpler case of unoccluded objects and then discuss the precise ambiguities and our proposed solution.

\paragraph{Special case: Unoccluded objects without contacts.}
Consider first the special case where each object is completely visible in each image, and does not make contact with any other part of the scene.
In this case, given a candidate object SDF, we can \emph{render} an object mask for each input viewpoint by aggregating the density along each ray.
We can then add a loss term that encourages this predicted mask to match the provided segmentation mask using a simple binary cross entropy loss.
Concretely, similar to Equation~\ref{render}, for every object and for every pixel we calculate a \emph{mask loss}:
\begin{align}
L_\textrm{mask} = \sum_k \sum_{\mathbf{j}}\text{BCE}(\hat{O}_k(\mathbf{j}) , M_k(\mathbf{j})) \label{eq:naivemask} \text{, } \hat{O}_k = \sum_i T_i \alpha_i^k
\end{align}
Here, $k$ ranges over objects, $\mathbf{j}$ is a pixel, $i$ ranges over samples along the ray, $\hat{O}_k$ is a score representing the opacity of this pixel in the $k$-th SDF, and $M_k$ is the ground-truth segmentation indicating if this pixel is part of the $k$-th object.

\begin{figure}[t]
\centering
\includegraphics[width=0.3\linewidth]{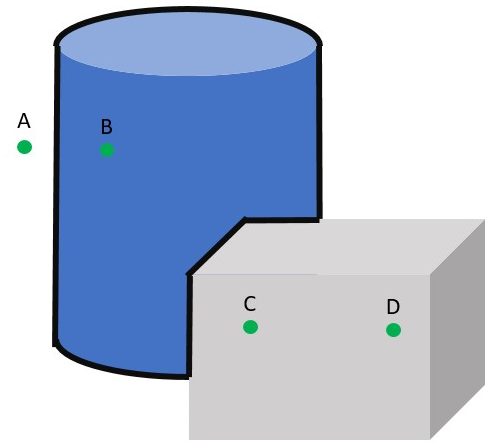}
\includegraphics[width=0.3\linewidth]{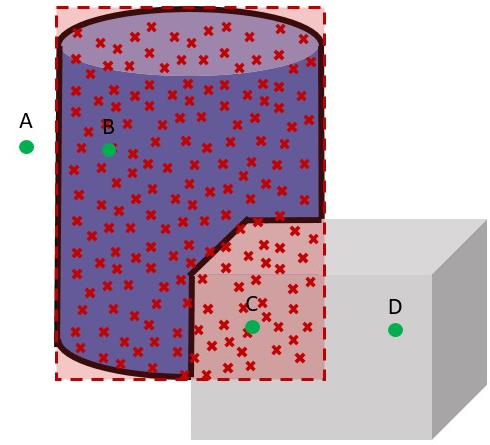}
\caption{An occlusion event. The object of interest is the blue cylinder. On the left is the segmentation mask. On the right, the crosses (not included in the segmentation mask) represent points on the blue object that are visible in other views but occluded in this view. The red dotted box is the amodal mask, and its intersection with the occluding cuboid is the set of pixels that are  ``present''  in the blue object, but occluded in this view.}
\label{fig:occlusion}
\end{figure}

This loss, as proposed in NeuS~\cite{neus}, causes problems in scenes with occlusions.
To see this, consider Figure~\ref{fig:occlusion}, where a blue object is occluded by a gray box. 
The segmentation only shows the visible parts of the blue object.
Clearly, point A must lie outside the blue object and thus the accumulated density along the corresponding ray will tend toward 0, as specified by the mask loss in Equation \ref{eq:naivemask}.
The mask loss also has the correct behavior for pixel B, which shows a point on the object surface and so the corresponding ray must have a high density point along it.
However, pixels C and D land on the gray box that occludes the object of interest, and so are not part of the mask.
The mask loss would suggest that the rays from these pixels should lie completely outside the object.
Clearly, this is not the right behavior at pixel C, and thus we need a different strategy to handle occlusion.
\paragraph{Resolving occlusion:}
One option is to not impose any loss on points C and D at all. 
In other words, we could exclude all pixels where the object of interest is occluded by another object.
Past work proposes an occlusion-aware loss which has a similar effect~\cite{objsdf++}.
However, the effect of this is that the trained SDF may now include artifacts that are occluded from view in all images without incurring any penalty.
While this kind of an object is \emph{possible} given the input views, our intuition tells us that it is highly unlikely.

Instead, we propose a prior that, to the extent possible, the object should only include surfaces that are visible from at least some input view.
In other words, we would like a \emph{compact completion} of the visible surfaces that we see in the input views.
Thus, in Figure~\ref{fig:occlusion}, we would be okay with the object including point C (because it is near projections of other surface points that are observed in other views), but not okay with any artifacts that include the point D.

We formalize this intuition as follows.
We backproject all annotated pixels for object $k$ from all input views into the reconstructed 3D scene to create a cloud of 3D points that are known to belong to this object, $\mathbf{p}_k$.
We then project all these points into every view without regard to occlusion (producing e.g., the crosses in Figure~\ref{fig:occlusion}).
In each view, we then take the \emph{bounding box} of these projected points; this is the \emph{amodal} bounding box of the object, $\mathcal{B}_k$ (the term \emph{amodal} completion  refers to the phenomenon where humans perceive the complete shape of a background object in spite of occlusion\cite{Amodal}). We then intersect this amodal bounding box with the provided segmentation masks of the \emph{other objects} to get a ``present-but-occluded'' mask $M^\textrm{occ}$.
We then only apply the mask loss above to pixels outside this present-but-occluded region.
\begin{align}
\mathcal{B}_k &= \text{Bounding Box} \left( \pi(\mathbf{p}_k) \right) \\
    M^\textrm{occ}_k &= \mathcal{B}_k \cap \left (\cup_{i\neq k} M_i \right )\\
    L_{\text{compactness}} &= \sum_k\sum_{\mathbf{j} \notin M^\textrm{occ}_k}\text{BCE}\left(M_k(\mathbf{j}), \hat{O}_k(\mathbf{j})\right)
\end{align}
Here $k$ indexes the objects.
Effectively, this compactness loss 
creates a compact region in 3D space (formed by the intersection of all the frusta corresponding to the amodal bounding boxes) in which the object is allowed to lie.
Any part of the object outside of this compact region is penalized irrespective of occlusion.

\paragraph{Resolving object interfaces.} A final step is to resolve object interfaces, to ensure that each object occupies a distinct region of 3D space and does not intersect others.
For this, we use a loss term, that we call the \emph{overlap} loss. It adds a penalty whenever the interiors of two objects overlap.
Concretely, suppose we have $K$ SDFs $f_1, \ldots, f_K$.
For every 3D point $\mathbf{p}$ sampled randomly in space, we identify the SDF that yields the most negative value (i.e., the object for which $\mathbf{p}$ is farthest into the interior), and penalize negative values from all other SDFs using a hinge loss:
\begin{align}
    k^* &= \arg\min_{k}\left(f_k(\mathbf{p})\right) \\
    L_{\text{overlap}}(\mathbf{p}) &= \sum_{k\neq k^*} \max\left(f_k(\mathbf{p}),0\right)
\end{align}

Our final loss function is:
\begin{equation}
    L = L_{\text{color}} + \lambda L_{\text{eik}} + \beta L_{\text{compactness}} + \gamma L_{\text{overlap}}.
\end{equation}
We train all $K$ SDFs in parallel using this loss. Where we set the hyper-parameters to be ($\lambda = 0.1$, $\beta =0.9,$  $\gamma = 0.001$). We tested on 1 RTX 3090 GPU. We used a batch size of 512 and 64 for the full scene reconstruction and per-object reconstruction respectively. 

\paragraph{Initialization.}
One challenge with the proposed approach is that it trains $K$ different SDFs, and so can be up to $K$ times as expensive as training the single scene SDF. 
Prior work uses various strategies to reduce this training cost, such as sharing layers between the SDFs~\cite{objsdf++} or distilling from the scene SDF~\cite{objcompnerf}.
We propose a simpler strategy that significantly reduces running time (to a few hours instead of days) and yet preserves details: we initialize each SDF with a copy of the full scene SDF (unlike ObjectSDF++, which uses a sphere initialization).
Because each SDF starts with geometry that matches the scene, it has all the details and matches the input images by default. All the network has to do is to ``cut off'' the scene SDF in the appropriate regions.

\section{Benchmark}
\label{sec:dataset}

\begin{figure}
\begin{center}
\includegraphics[width=0.7\linewidth]{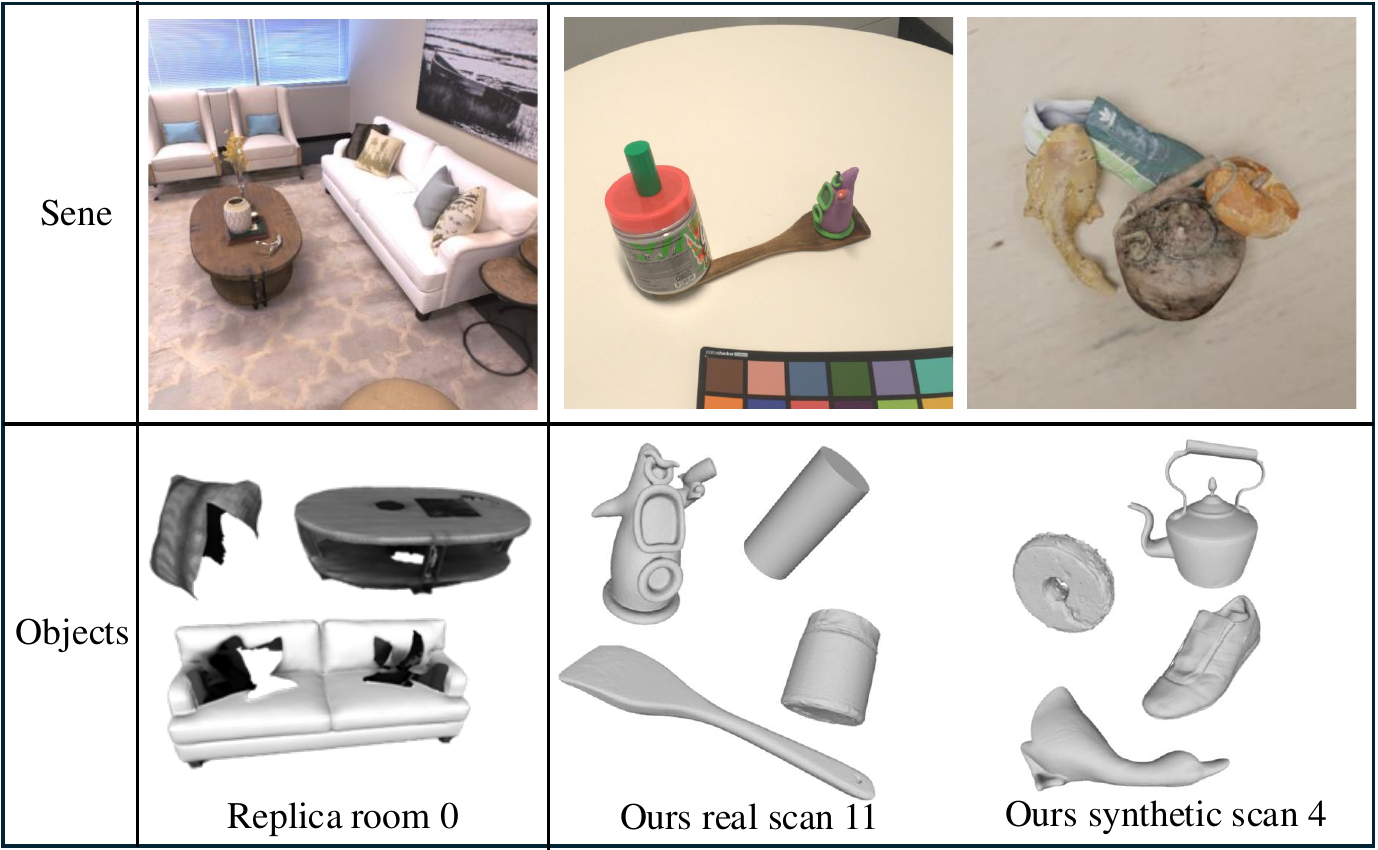}
\caption{Left: Previous datasets, like Replica, feature objects that only includes visible surfaces, not complete surfaces (including hidden surfaces). As a result, using the cropped sub-meshes as ground-truth for object separation is not an adequate evaluation. Middle and right: Our proposed dataset with complete individual objects.}
\label{fig:replica}
\end{center}
\end{figure}
Previous scene decomposition techniques evaluate their methods on benchmark datasets like Replica and ScanNet. A critical limitation of these is that they do not offer complete ground truth geometries for the reconstructed objects. More specifically, per-object meshes are extracted from the full ground truth mesh of the indoor scene by cropping the ground truth mesh with the semantic masks and therefore lack completeness in regions occluded by other objects (Figure~\ref{fig:replica}). 
We introduce a new benchmark for 3D scene decomposition techniques, consisting of 30 real-world scenes and 5 synthetically generated ones. The scenes contains different combinations of objects in close contact, and we provide a high-quality complete mesh of each object.

\subsection{Dataset}
\noindent \textbf{Real-World Scenes.} 
We provide 22 individual 3D scanned objects and 32 scenes, each created using a combination of the individual objects. To scan the individual objects, we use Polycam \cite{Polycam} (for analysis of Polycam, please refer to the supplementary materials). The scenes are captured as raw images using a phone camera at a resolution of 3008 × 3008. We provide camera pose estimates from COLMAP \cite{Schonberger_2016_CVPR}, ground-truth meshes from Polycam \cite{Polycam}, and masks generated using our mask propagation strategy. Last, we provide rotations and translations that align the ground-truth object meshes to the objects in the scenes. 

\medskip
\noindent \textbf{Synthetic Scenes.} We provide 5 synthetic scenes composed by combining objects with varying geometric complexities. We used Blender \cite{Blender} to create the dataset, with each scene centered at the origin. We used white indoor scene environment lighting. We rendered the scenes with 500 samples at a resolution of 512×512 using the Cycles renderer, capturing 100 images from cameras positioned on the upper hemisphere around the subject. In addition to the multi-view images, we provide ground-truth poses, geometries, masks and transformations that align object meshes to the corresponding scene. Please refer to our supplementary material for more details on the creation of real and synthetic dataset. 

\subsection{Evaluation}
To evaluate our method and the baselines we show quantitative and qualitative results on our synthetic and real world benchmark datasets. For quantitative evaluation we report the precision and completion ratio. Precision is the ratio of reconstructed points that are within a distance of \bh{$\theta$} from the ground truth, and penalizes floaters.
Completion ratio is the ratio of ground truth points that are within a distance of \bh{$\theta$} of the reconstruction, and penalizes incomplete reconstructions.

The two-way Chamfer distance is also measured between evenly-sampled vertex points on the ground truth mesh and sampled vertex points on the predicted mesh obtained by running marching cubes on the trained SDF.

To calculate these metrics between predicted and ground-truth meshes, it's crucial to maintain similar point densities to prevent imbalances. This can be difficult if the two meshes are widely different in size. ObjectSDF++\cite{objsdf++} addresses this by clipping predicted meshes using ground-truth bounds to improve density similarity and remove outliers. However, this approach may artificially inflate precision by not penalizing for floaters outside the bounding box. Instead, we keep the meshes as is but propose a refinement technique that uses rejection sampling to maintain consistent point densities, adjusting for mesh saturation until the surface can't hold more points and ensuring a fair comparison. Please see the supplement for more details.

\section{Experiments}
We first evaluate how our mask propagation strategy performs with increasing number of anchors and iterations.
Then, we compare our full pipeline qualitatively (Fig.~\ref{fig:qlt_eval}) and quantitatively (Table.~\ref{tab:qnt_eval}) against two baselines, ObjectSDF++ \cite{objsdf++} and RICO \cite{li2023rico}. We benchmark these methods on the five synthetic datasets. Because ObjectSDF++ and RICO fail to produce meshes for some of the real-world scans, we evaluate on a  subset of 11 real scans for which all methods can produce valid meshes.
Finally, we ablate components of our proposed method to see their impact on the quality of our solution.

\subsection{Mask Propagation Evaluation}
We first evaluate the performance of our segmentation propagation approach. 
To do so, we use our synthetic dataset where all scenes have corresponding ground-truth segmentation.

The first column of each scan in Table \ref{tab:maskiou} shows the mIOU (Mean Intersection over union) for each iteration, starting with one anchor image. We observe that the first iteration generally performs poorly but the mIOU improves with more iterations; however, after the second iteration, the improvement becomes minimal. Our method can also take multiple anchor images if provided, this can be useful for example if all the objects are not visible in one image alone or the user wanted to provide more information. We evaluate the effect of providing multiple anchor masks in the second and third columns of each scan. However, after the third iteration, whether we start from a single image or multiple anchor masks, all converge to similar results, as shown in the third row.

\begin{figure}[t]
  \centering
  
   \includegraphics[width=0.9\linewidth]{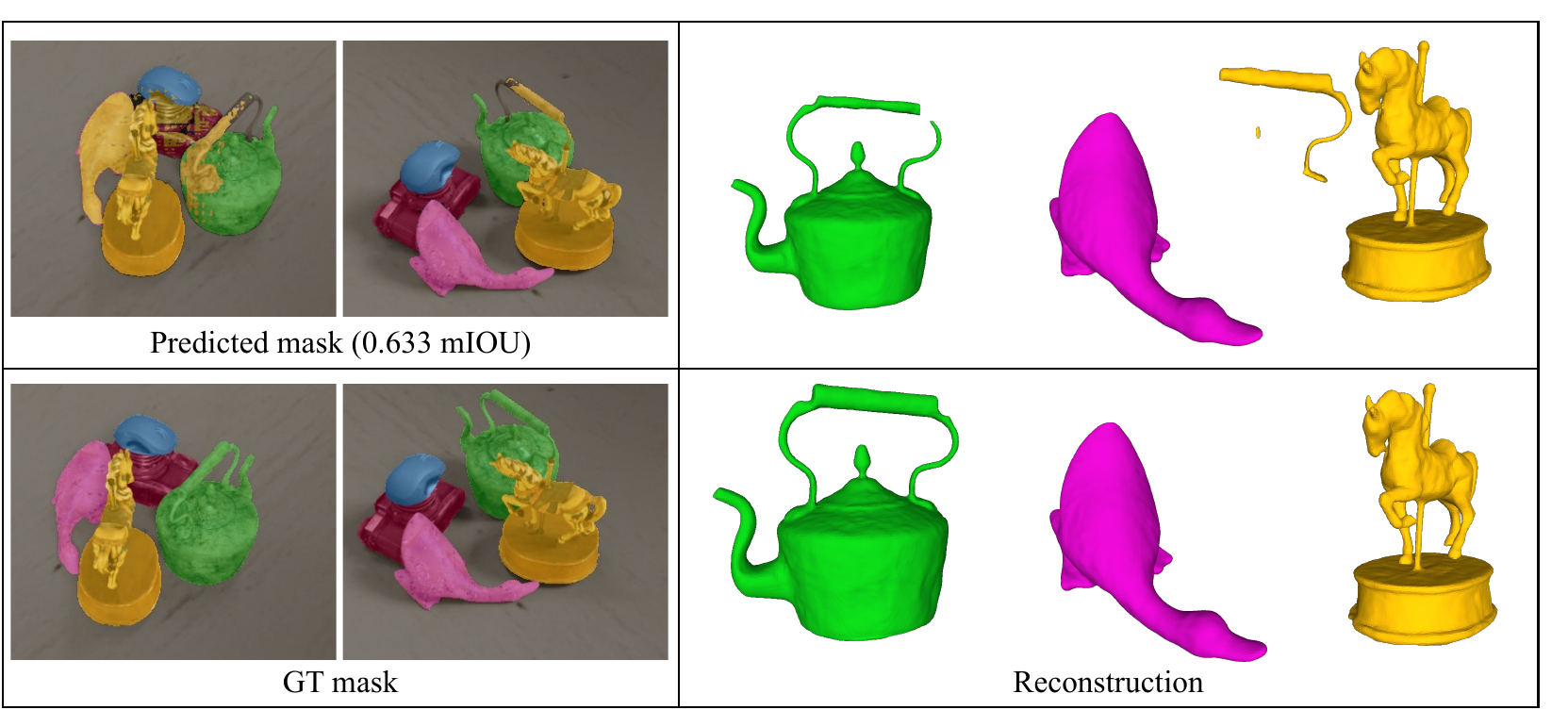} 
   \caption{Impact of Low mIOU (scan2 Table \ref{tab:maskiou}) on reconstruction. Top: An example of a low mIOU (0.633) mask, where the handle of the kettle and parts of the duck are sometimes misclassified as part of the horse. Despite this, our object separation pipeline remains robust. For instance, the duck has an accurate 3D mesh due to our scene initialization technique, and the mask labels are correct in other images. However, the 3D reconstruction of the horse includes part of the kettle handle, as most of the mask incorrectly classifies the handle as part of the horse. Bottom: Ground truth mask and its respective reconstruction.
   }
   
   \label{fig:mask_effect}
\end{figure}

A failure case of the mask propagation is presented in scan 2 in Table \ref{tab:maskiou}, where the mIOU is low. This is because some parts of one object end up being labeled as another object; for example, the duck in Figure \ref{fig:mask_effect} (top left) is classified as the horse. 
Note that the same surface may be correctly labeled in a different image, since SAM is performed independently for each image.

One may ask how this failure impacts our final reconstruction and separation result.  Figure \ref{fig:mask_effect} shows, despite the low mIOU mask of scan 2, our object separation module still reconstructs plausible results. This is due to the majority of the masks are still correctly labeled.

\begin{table}[h!]
  \begin{center}
    \caption{We report the mIOU values of the predicted masks using our mask propagation strategy, varying the number of propagation iterations and anchor images per each scan. Mask quality does not improve much after the second round of mask propagation and adding additional anchors does not offer much improvement after a few propagation iterations.}
    \renewcommand{\arraystretch}{1.3}
    \resizebox{\columnwidth}{!}{%
      \begin{tabular}{c c c c c c c c c c c c c c c c c c c c c c}
      \toprule
        \multicolumn{2}{c}{Scan} & \multicolumn{3}{c}{1} & &\multicolumn{3}{c}{2} & &\multicolumn{3}{c}{3} & &\multicolumn{3}{c}{4} & &\multicolumn{3}{c}{5} \\
    \hline
        \multicolumn{2}{c}{No of anchors} & 1 & 2 & 3 && 1 & 2 & 3 && 1 & 2 & 3 && 1 & 2 & 3 && 1 & 2 & 3\\
        \hline
        \multirow{3}{*}{iter} 
        & 1 & 0.73 & 0.77 & 0.86 && 0.60 & 0.63 & 0.65 && 0.71 & 0.71 & 0.77 && 0.86 & 0.91 & 0.91 && 0.80 & 0.77 & 0.90 \\ 
 & 2 & 0.90 & 0.90 & 0.91 && 0.64 & 0.64 & 0.64 && 0.78 & 0.78 & 0.78 && 0.91 & 0.92 & 0.91 && 0.94 & 0.94 & 0.94 \\ 
 & 3 & 0.91 & 0.91 & 0.91 && 0.63 & 0.63 & 0.63 && 0.78 & 0.78 & 0.78 && 0.90 & 0.91 & 0.91 && 0.94 & 0.94 & 0.94 \\ 
        \toprule
      \end{tabular}
    }
    \label{tab:maskiou}
  \end{center}
\end{table}

\vspace{4mm}
\subsection{Reconstruction Evaluation}
\begin{table}[h]
\centering
\scriptsize
\caption{\textbf{Quantitative evaluation: } RICO performs the lowest among all methods. ObjectSDF++ performs well on synthetic data, but its performance drops on real data, especially in terms of precision ratio. This drop is due to the imperfect masks in the real scans. On both synthetic and real datasets, our method outperforms the baseline in all metrics. We used GT masks for the synthetic evaluation and masks generated by our mask propagation for the real dataset.} 
\setlength{\tabcolsep}{1.7mm}
\begin{tabular}{c c c c c }
\toprule
 Dataset & Metrics & RICO & ObjectSDF++ & Ours \\ 
 \toprule
\multirow{3}{*}{Synthetic} & Chamfer ↓ & 0.374 & 0.018 & 0.010 \\ 
 & Prec. Ratio ↑ & 0.449 & 0.937 & 0.960 \\ 
 & Comp. Ratio ↑ & 0.868 & 0.946 & 0.991 \\ 
 \hline \rule{0pt}{2.5ex}
\multirow{3}{*}{Real} & Chamfer ↓ & 0.138 & 0.096 & 0.015 \\ 
 & Prec. Ratio ↑ & 0.537 & 0.763 & 0.921 \\ 
 & Comp. Ratio ↑ & 0.595 & 0.835 & 0.927 \\ 
 \toprule
\end{tabular}
\label{tab:qnt_eval}
\end{table}

\begin{figure}
  \centering
   \includegraphics[width=0.5\textwidth]{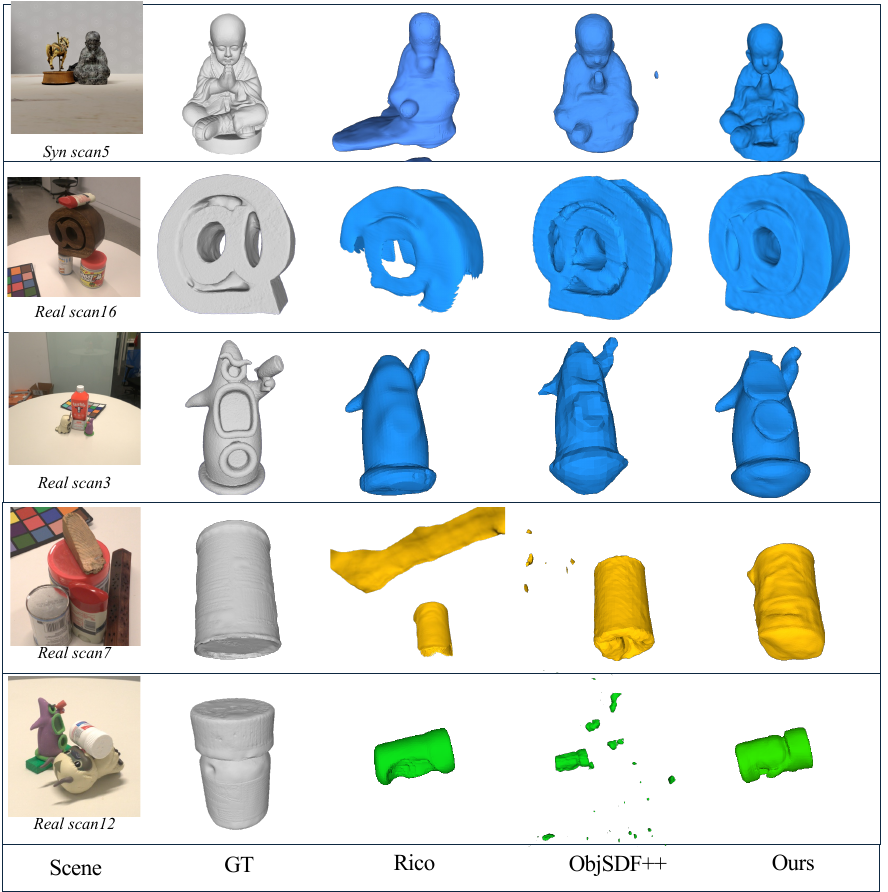} %

   \caption{\textbf{Qualitative Comparison}: RICO and ObjectSDF++ produce floating artifacts, as shown in Real scans 7 and 12. RICO also sometimes carves out the object, leaving a hollow area, as shown in Real scan 16, 3 and 12. In contrast, our method produces fewer artifacts while also providing more detail.
}
   \label{fig:qlt_eval}
\end{figure}

Table~\ref{tab:qnt_eval} results reveal the failures of the baselines. First, RICO performs poorly in the quantitative results, despite having decent qualitative results. The reason is that RICO, while often complete, produces huge floaters like the one visualized in `Real scan7' in Figure~\ref{fig:qlt_eval} which significantly hurts the quantitative performance. While RICO achieves good completion metrics, it struggles to precisely generate meshes of the object of interest.

Second, ObjectSDF++, while competitive on the synthetic datasets, loses out in the real dataset benchmarks. Unlike the synthetic ground-truth masks, the masks used in the real-world benchmark of ObjectSDF++ were obtained using our proposed mask propagation strategy, which is still imperfect. This not only results in floaters, which are not handled due to the absence of a compactness loss, but also a loss of detail of objects at sharp edges as shown in `Real scan3' and `Real scan16'. In contrast, we initialize the object SDF from the reconstructed scene, resulting in more robust results.

From Figures \ref{fig:qlt_eval} and Table \ref{tab:qnt_eval}, we can conclude our method produces results with higher quality and fewer floating artifacts.  Most of our quantitative improvement comes from the lack of undesired artifacts like floaters and carved holes. The remaining improvement, more evident qualitatively, comes from the scene initialization. 

\begin{figure}[t]
  \centering
   \includegraphics[width=1\linewidth]{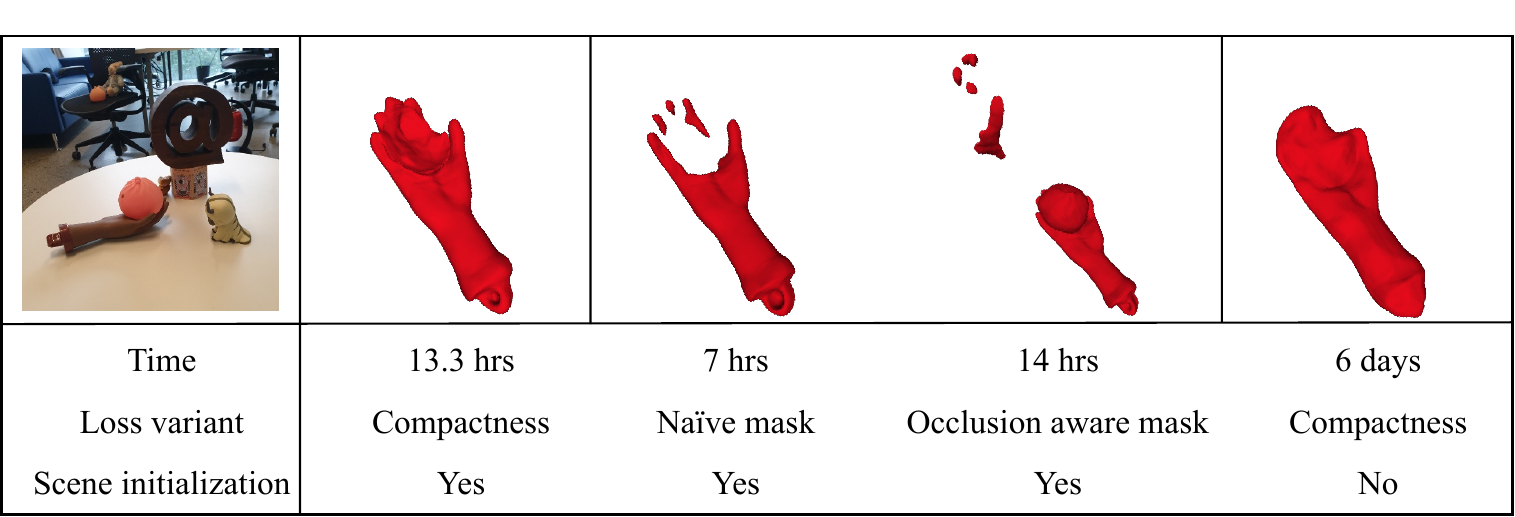} %
   \caption{\textbf{Importance of the compactness loss and initialization.} Left: our compactness loss and initialization together avoids floating artifact and achieves high-quality results. Middle: a naive mask loss as in NeuS carves out objects whenever there is an occlusion and with occlusion aware mask, and we see floating artifacts in unobserved parts of the scene. Right: without scene initialization details are lost and the runtime grows significantly. 
   }
   
   \label{fig:ablation-loss}
\end{figure}
\subsection{Ablation}
To understand the importance of our contributions, we ablate the proposed compactness loss and the scene initialization as shown in Figure~\ref{fig:ablation-loss}.

To evaluate the compactness loss, we compare to two alternatives:
\begin{enumerate}
\item The first baseline is the naive mask loss used in NeuS, which does not take object occlusion into account. This loss is defined in Equation~\eqref{eq:naivemask}. 
\item The second alternative without compactness is an \emph{occlusion-aware mask loss}: we simply apply the mask loss only to the unoccluded pixels, i.e., to discount pixels that are marked as belonging to other objects. 
\begin{align}
\tilde{M}^\textrm{occ}_k &=  \left (\cup_{i\neq k} M_i \right )\\
L_{\text{occ-aware}} &= \sum_k\sum_{\mathbf{j} \notin \tilde{M}^\textrm{occ}_k}\text{BCE}\left(M_k(\mathbf{j}), \hat{O}_k(\mathbf{j})\right)
\end{align}
While this strategy correctly avoids penalizing pixels that are part of the object but occluded, it does not encourage the object to be compact. As such, the model is free to hallucinate other floaters as long as they are completely occluded in the view. 
\end{enumerate}

The three loss variants are shown in the first three columns of Figure~\ref{fig:ablation-loss}. When using naive mask loss, object geometries are carved out, resulting in incomplete reconstructions. This is because the model is penalized whenever it produces a surface that is occluded, evident in the hand's fingers becoming detached as a result of the object sitting on it. The occlusion-aware mask loss prevents the objects from being carved out, but introduces floaters, sometimes \emph{inside} the other objects, which are reconstructed as hollow shells. This occurs because any floater that is completely inside the shell of another object will never be visible and therefore never be penalized. 
The compactness loss both removes floaters and prevents the objects from being carved out.

The last column of Figure~\ref{fig:ablation-loss} shows the reconstruction without the scene initialization. In this case, the reconstruction quality is significantly reduced and reconstruction requires a prohibitively long time. Scene initialization is critical to reducing computational time and obtaining high-quality results.

    \section{Conclusion}
We proposed \methodname, a method that separates objects in a scene into individual high-resolution meshes by automatically generating segmentation masks for all multi-view training images from a few clicks on just one image. We introduced compactness loss, a novel loss function that removes many of the floaters that have plagued prior methods. Finally, we show that initializing the per-object models with the scene model not only improves convergence and reduces training time but also maintains the details of the objects.

\renewcommand{\thesection}{\Alph{section}}
\setcounter{section}{0}

\title{
\bfseries\fontsize{14}{16}\selectfont Supplementary Material}

\maketitle
\section{Runtime analysis}
We provide a runtime analysis for all stages of our method, considering an image size of 512×512 with a single RTX 3090 GPU.

\begin{itemize}
    \item Stage 1: Scene reconstruction for 200k iterations takes 5.8 hours.
    \item Stage 2: Segmentation takes 2 minute per image.
    \item Stage 3: The amount of time Object Separation takes depends on the number of objects in the scene. For two, four, and six objects, Object Separation takes 2.7, 3.5, and 7.5 hours respectively.

\end{itemize}
\section{Dataset}
The problem of object separation in 3D reconstruction is a fairly new topic and, as such, lacks the proper benchmark dataset. Previous methods evaluated their approach on a cropped sub-meshes from a full scene, which has in holes in occluded regions. Thus, during the evaluation, the area that needs to be properly evaluated will be ignored. Figure \ref{fig:replica} illustrates the issue with previous datasets. To address the gap in the literature, we introduce a new benchmarking dataset composed of real-world and synthetic scenes. Below, are the details on how we created the dataset.
\begin{figure}[t]
  \centering
   \includegraphics[width=1\linewidth]{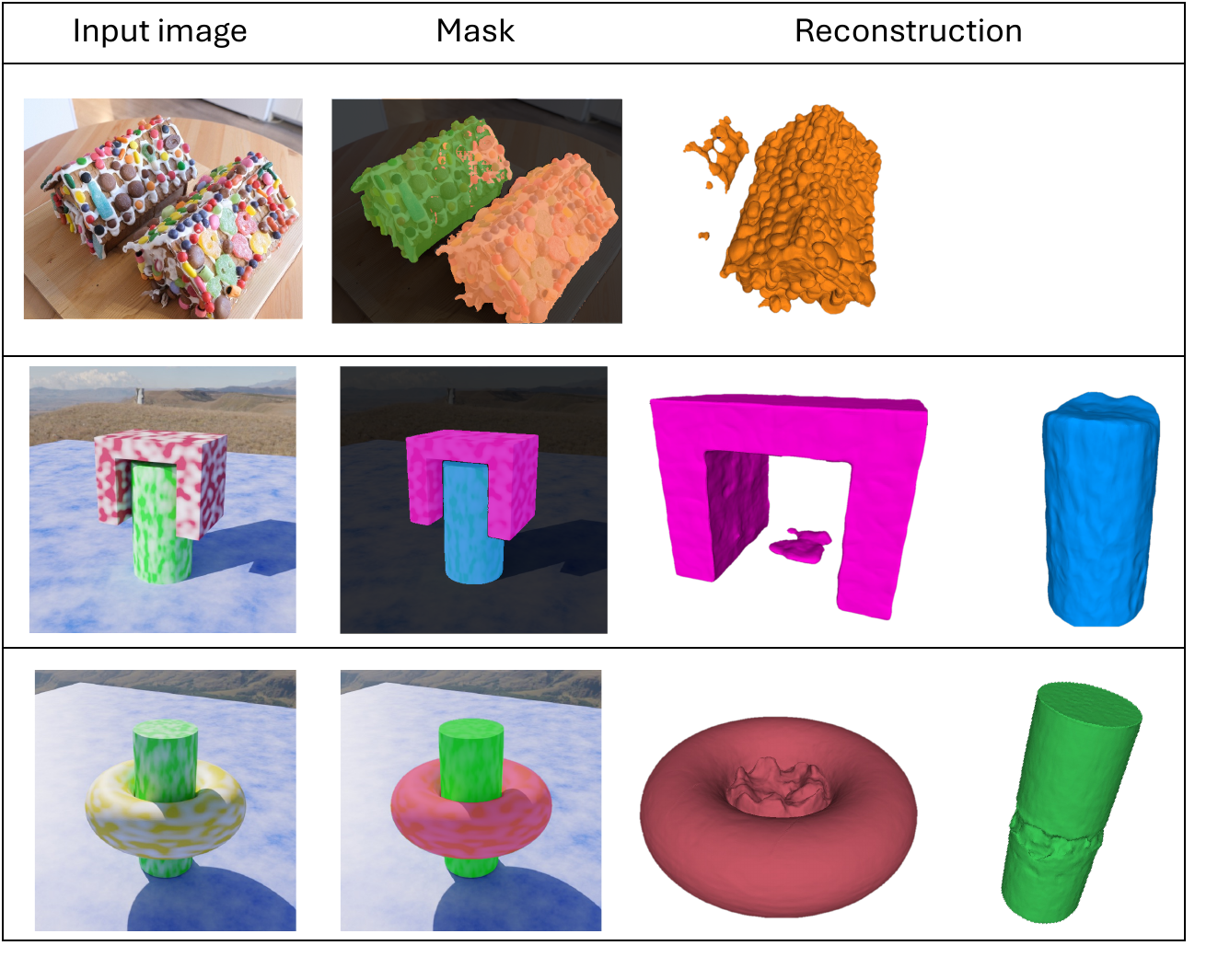} %
   \caption{[first row] illustrates our method's failure due to segmentation errors. Here, SAM itself struggles to obtain the correct mask as the two gingerbread houses are colorful, making it difficult to segment them. In the [second row], the segmentation is correct, but there is a flat surface floating in the pink box. This is due to the limitation of our amodal bounding box, which is contain this space from all views. In addition the overlap loss may not be computed at area as we are using  fixed random points (of 10,000) in 3D space to compute the overlap loss and the pink floating surface is thin, and the points may not lie within this area. In the [third row], both the torus and the pipe occupy the space. Once again, this highlights the shortcomings of the amodal bounding box and overlap loss.
   }
   \label{fig:limitation}
\end{figure}
\subsection{Real-world dataset creation}
Figure \ref{fig:capture_method} shows the steps we took in creating the dataset. First, we scanned the individual object to obtain the ground truth mesh using Polycam. Second, we captured the scene using a handheld camera and then obtained the camera pose using COLMAP. The fourth step involved obtaining the full scene reconstruction so that we could use it to align the individual meshes. We trained vanilla NeuS to obtain the full mesh. The fifth step involved aligning the meshes, We used Blender software to transform the individual meshes to their respective positions within the scene. To further refine the alignment and avoid human error, we applied ICP to align the meshes together. Ultimately, we obtained each scene images, camera pose, and the transformation matrix for each individual object. Below, we describe in more detail how we scanned the individual meshes and the scene.
\subsubsection{Individual mesh scanning.} 
\begin{figure}[t]
  \centering
   \includegraphics[width=1\linewidth]{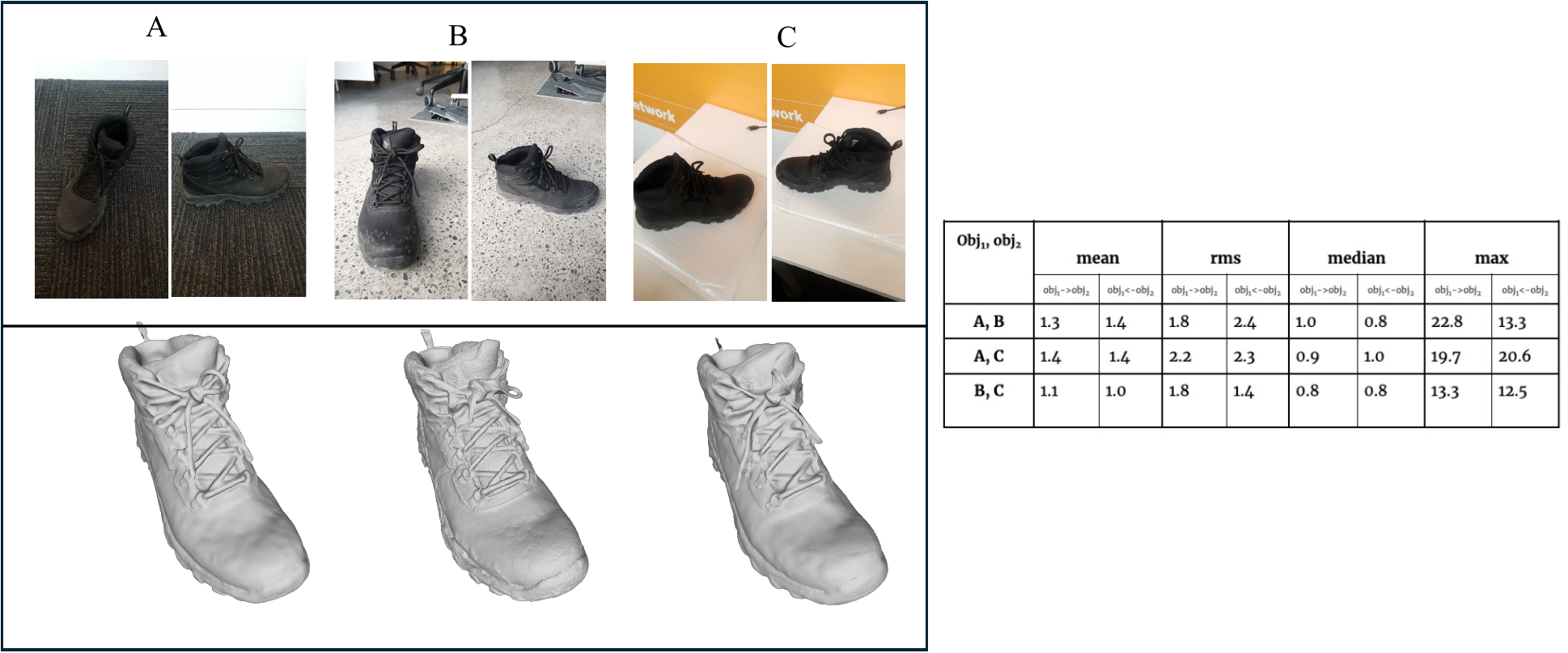} %
   \caption{ Analysis of the robustness of Polycam: [Left] images captured in different environments and their respective Polycam meshes. [Right] Quantitative evaluation where we compare each mesh against each other (unit in millimeters). We can observe that both qualitatively and quantitatively, Polycam is consistent. 
   }
   \label{fig:analysis}
\end{figure}

We provide 22 object scans. We collected 80-150 images per object and used Polycam to generate the full mesh. We set the option \textit{Isolate object from environment} to true and exported the final mesh in raw format. Since Polycam is not open-source software, we conducted an analysis of its reconstruction quality. We performed an experiment where we used the same object in different environments. We captured the object and obtained a mesh with Polycam and analyzed the robustness of Polycam. Figure \ref{fig:analysis} illustrates our analysis. We used a boot and took pictures in three different environments and used these three sets of images to obtain the meshes. Then we compared them to each other to see if there is a significant quality drop. However, we found that they are very similar, and we concluded that Polycam is consistent in its output.
\subsubsection{Scene capture.} 
\begin{figure}[t]
  \centering
   \includegraphics[width=1\linewidth]{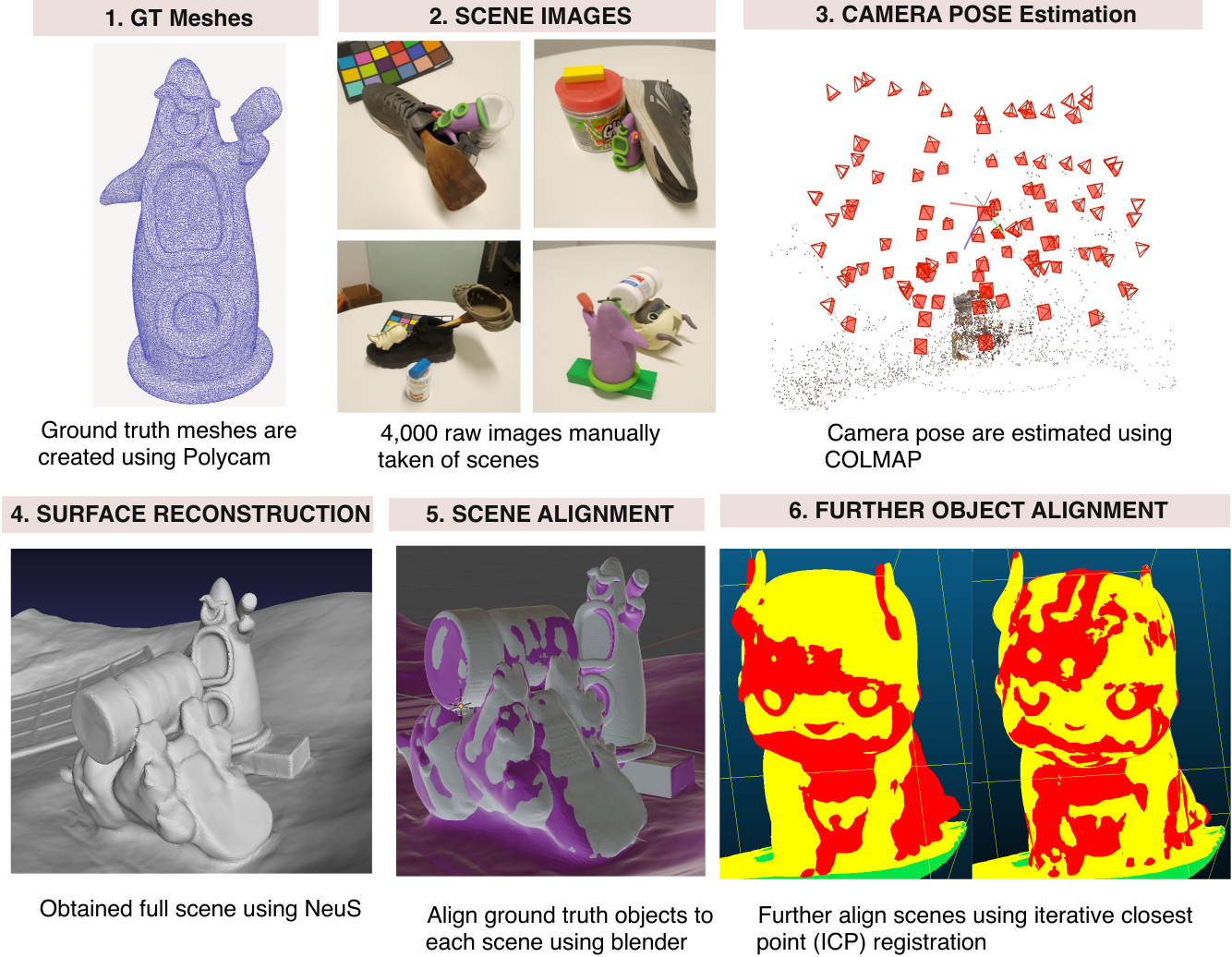} %
   \caption{\textbf{Real-world dataset creation steps}: (1) Scanning individual objects with Polycam for ground truth meshes, (2) Capturing the scene with a handheld camera, (3) Estimating camera pose with COLMAP, (4) Full scene reconstruction for mesh alignment, (5) Aligning individual meshes with Blender to the full scene, and (6) Further aligning the meshes with ICP for precision. From this process, we obtain scene images, camera poses, and transformation matrices for each object.
   }
   \label{fig:capture_method}
\end{figure}

We provide 32 real-world scene. We used a Samsung Note9 with a 12-megapixel camera to capture the scenes, utilizing the raw option in the pro-mode to capture raw images. The original images are $3008 \times 3008$ pixels with 16-bit depth. We converted the raw format to .png for lossless compression and downscaled it to $1002 \times 1002$ pixels. We show the steps we took to collect the dataset on Figure \ref{fig:capture_method}. Both the individual scene and individual objects can be found in Figure \ref{fig:scenes}.

\begin{figure*}
  \centering
   \includegraphics[width=1\linewidth]{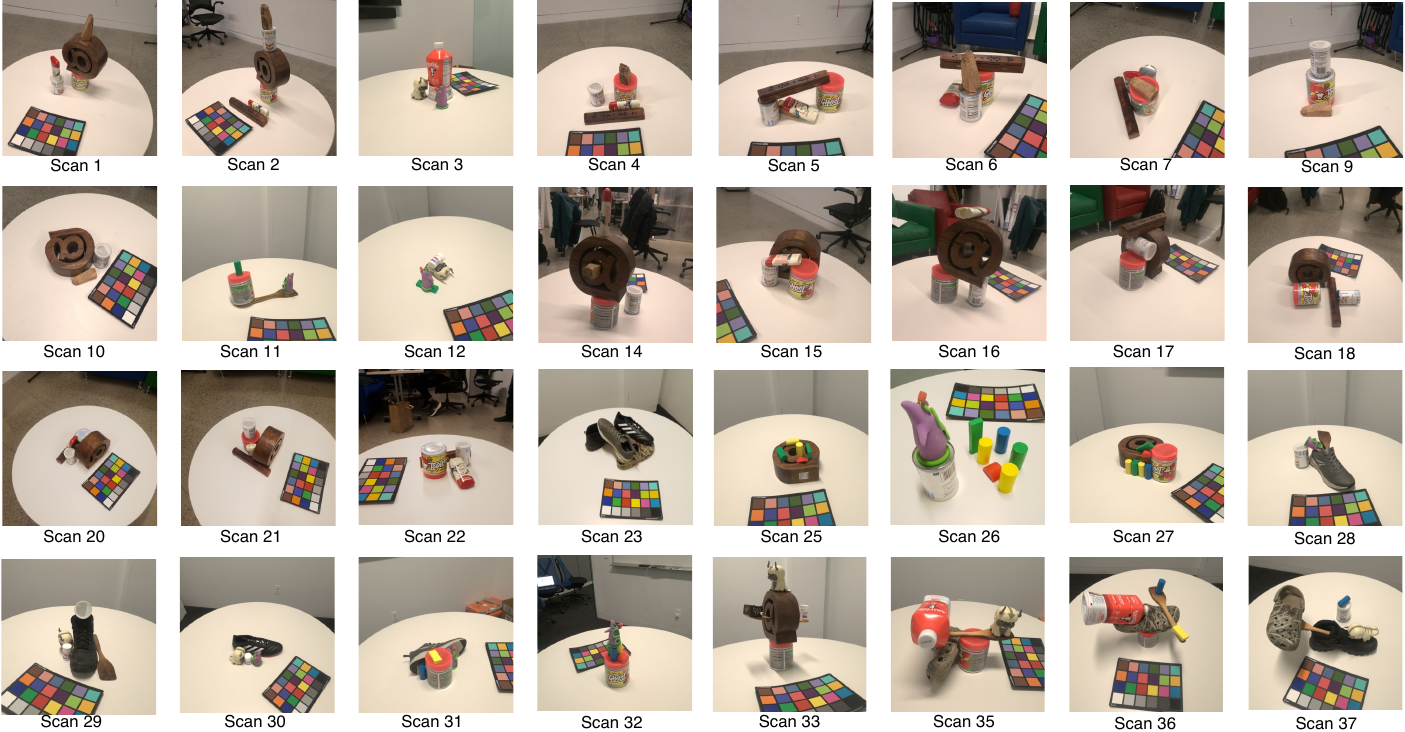} %
   \caption{Full list of captured real-world scenes. 
   }
   \label{fig:scenes}
\end{figure*}

\begin{figure*}
  \centering
   \includegraphics[width=0.8\linewidth]{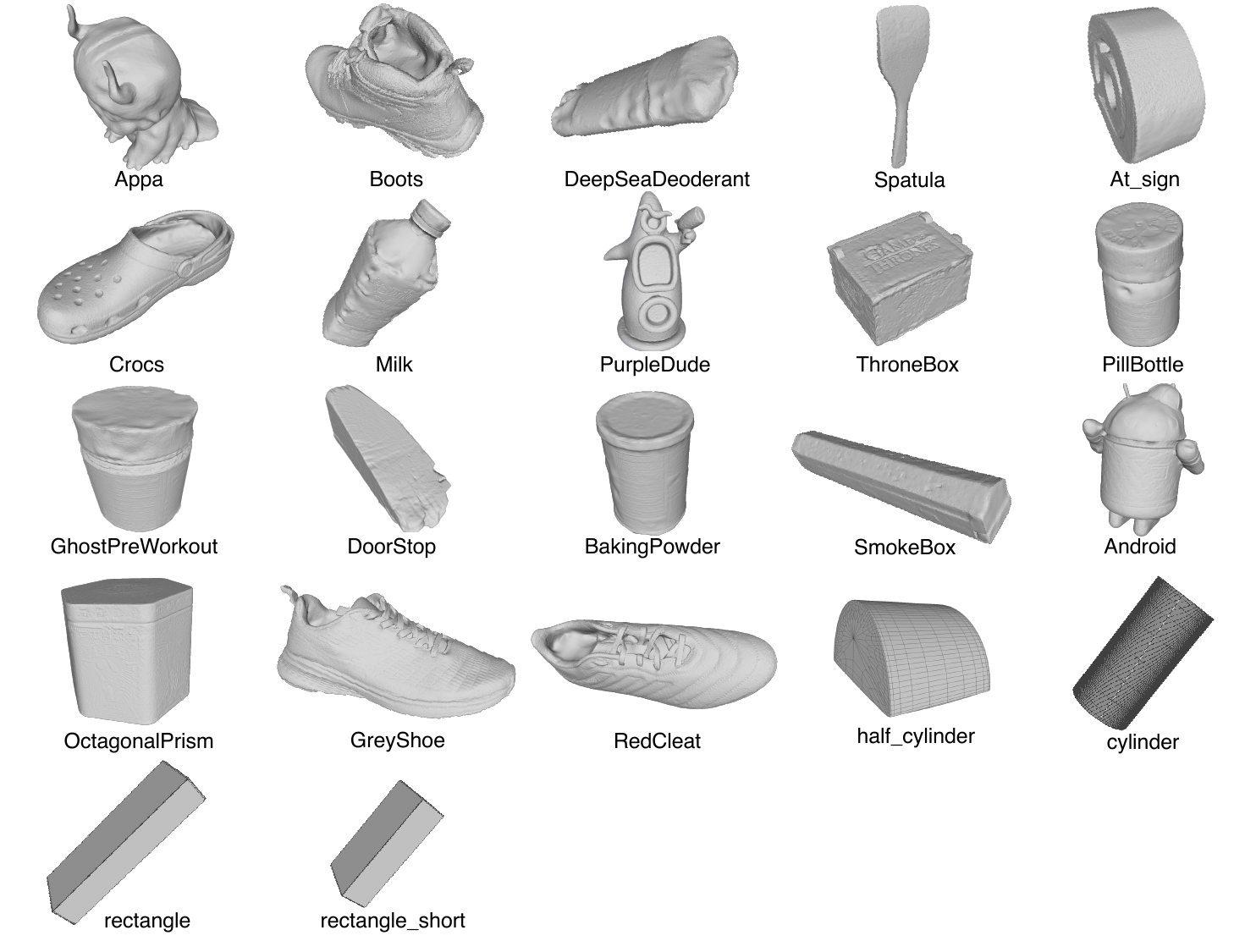} %
   \caption{Full list of scanned individual meshes using Polycam.
   }
   \label{fig:scenes}
\end{figure*}
\begin{figure*}
  \centering
  
   \includegraphics[width=0.7\linewidth]{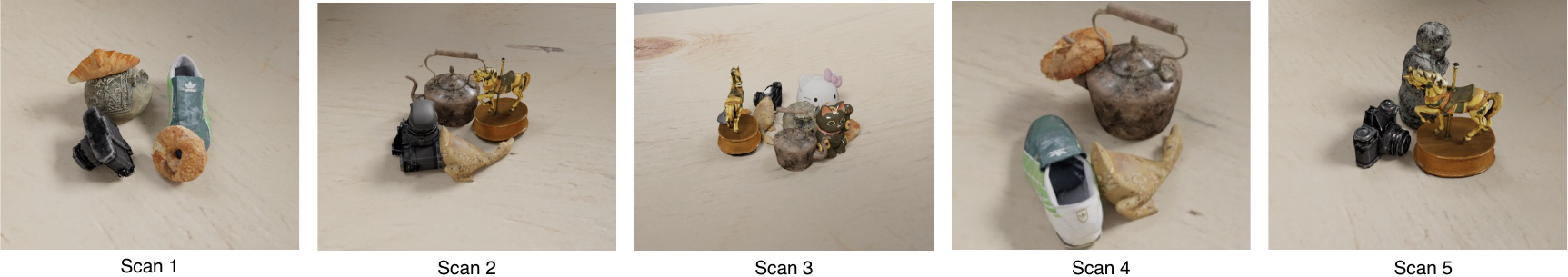} %
   \caption{Full list of synthetic scenes.
   }
   \label{fig:scenes}
\end{figure*}

\subsection{Synthetic data creation}

We generated five realistic scenes, each with its own level of difficulty. Each scene has $N$ objects, $N$ ranging from 3 to 10. We used Blender to create the dataset, with each scene centered at the origin. We used white indoor scene environment lighting. We rendered the scenes with 500 samples at a resolution of $512\times 512$ using the Cycles renderer, capturing 100 images from cameras positioned on the upper hemisphere around the subject.

\section{Rejection sampling}
When calculating the point-to-point Chamfer distance between predicted and ground-truth meshes, it is important to ensure locally similar point densities to avoid one of the directional Chamfer distances from being larger than the other. While the ideal metric here is the point-to-surface Chamfer distances (i.e. the average unsigned distances), this is often prohibitively slow; hence, it is common practice to resample the two point clouds to have the same number of points for an accurate point-to-point Chamfer distance. However, when the predicted mesh contains floaters and extraneous artifacts, this results in a diluted sampled point cloud and causes the densities to differ in the region of interest. The reverse holds true when the predicted mesh experiences carved out regions that lower its surface area for point sampling. ObjectSDF++ and Rico are evaluated on Replica and ScanNet, and they clip the predicted meshes using the 3D bounds calculated from the ground-truth meshes. While this was likely done to try to ensure similar point densities, it removes any floater artifacts that exist outside the clipping bounds, yielding in a artificially lower smaller precision metric. We revise this evaluation by using a rejection sampling based approach that samples points only if it is some radius away from the growing list of samples. If the desired number of points is large enough to saturate the mesh (i.e. desired number of points is impossible due to the radius constraint), we ensure that the point density is similar between the two meshes.
\section{Ablation}
\subsection{Effect of mask propagation on object reconstruction:} 
\begin{figure}[t]
  \centering
   \includegraphics[width=0.7\linewidth]{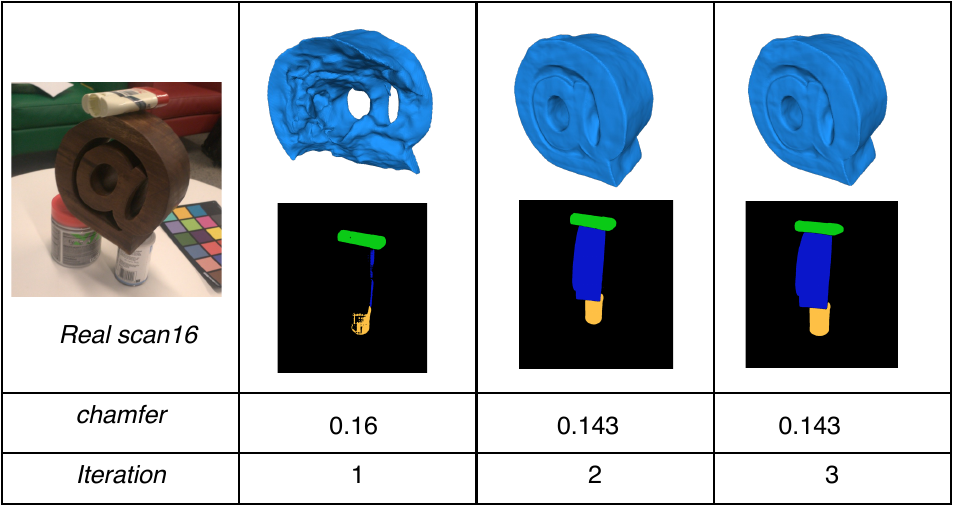} %
   \caption{Effect of mask propagation on the reconstruction, [Top] reconstruction of using the mask, [bottom] mask used for.
   }
   \label{fig:mask_ablation}
\end{figure}

Figure \ref{fig:mask_ablation} shows the effect of mask propagation using the mask obtained from each iteration. We see that the first iteration is not enough to capture the full object, as some parts of the segmentation are missing, resulting in carving out. However, after the second iteration, we observe that we can obtain the full geometry.

\subsection{Effect of increasing number of SDF parameters of baselines}
One possible reason why {\methodname} achieves more details in the reconstruction is that increase in model capacity with a whole SDF network being dedicated to each model. On the other hand, ObjectSDF++ and RICO use a single SDF network backbone with separate heads for each object's SDF. In order to determine if the expressivity of the SDF networks in RICO and ObjectSDF++ is the limiting factor, we increase the learnable parameters of the SDF networks in RICO and ObjectSDF++ by the number of objects $k$. For ObjectSDF++, we increase the dimensionality of the feature vector learned at each level of the hash-grid by $k$. For RICO, we increase the width of the network by $\lceil\sqrt{k}\rceil$. These modified models are called RICO* and ObjSDF++*.

\begin{figure*}[t]
  \centering
   \includegraphics[width=\linewidth]{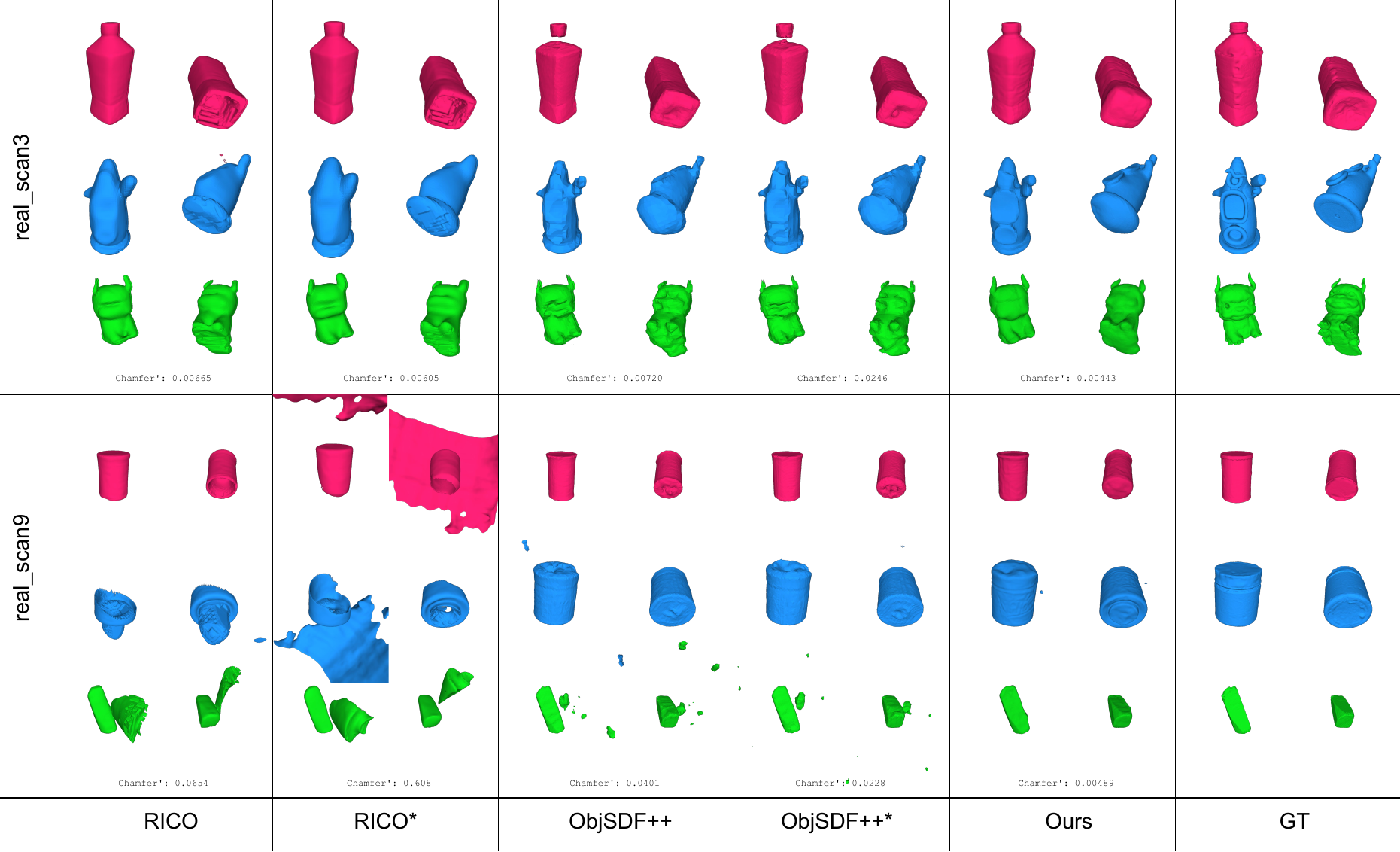} %
   \caption{Comparison of increasing the expressivity of the model backbone of RICO and ObjectSDF++
   }
   \label{fig:apples2apples}
\end{figure*}

Figure ~\ref{fig:apples2apples} shows the evaluation results of this comparison. RICO* tends to over-smooth the geometry and create more floater artifacts than RICO. ObjectSDF++ achieves better reconstruction quality and even reduces the number of floaters. However, {\methodname} still achieves the best qualitative and quantitative results among the baselines, demonstrating the effectiveness of scene initialization for learning the geometries with more parameters.

\begin{figure*}[t]
  \centering
   \includegraphics[width=\linewidth]{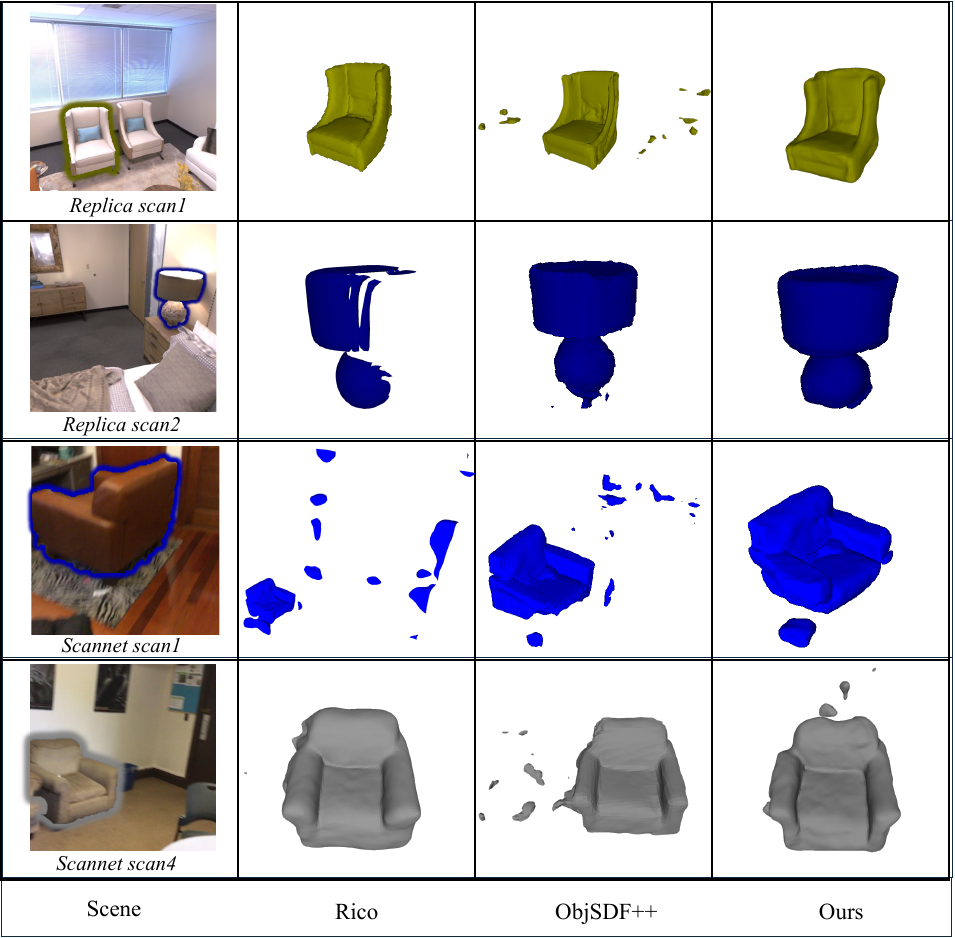} %
   \caption{Qualitative comparison on large indoor scenes: Due to our compactness loss, our method results in fewer artifacts compared to the baseline, which is plagued by floating artifacts, most apparent in row 3, and carving of the objects shown in row 2 of the RICO output.
   }
   \label{fig:replica_scannet}
\end{figure*}
\section{Additional Results}
\begin{figure*}[!b]
  \centering
   \includegraphics[width=1\textwidth]{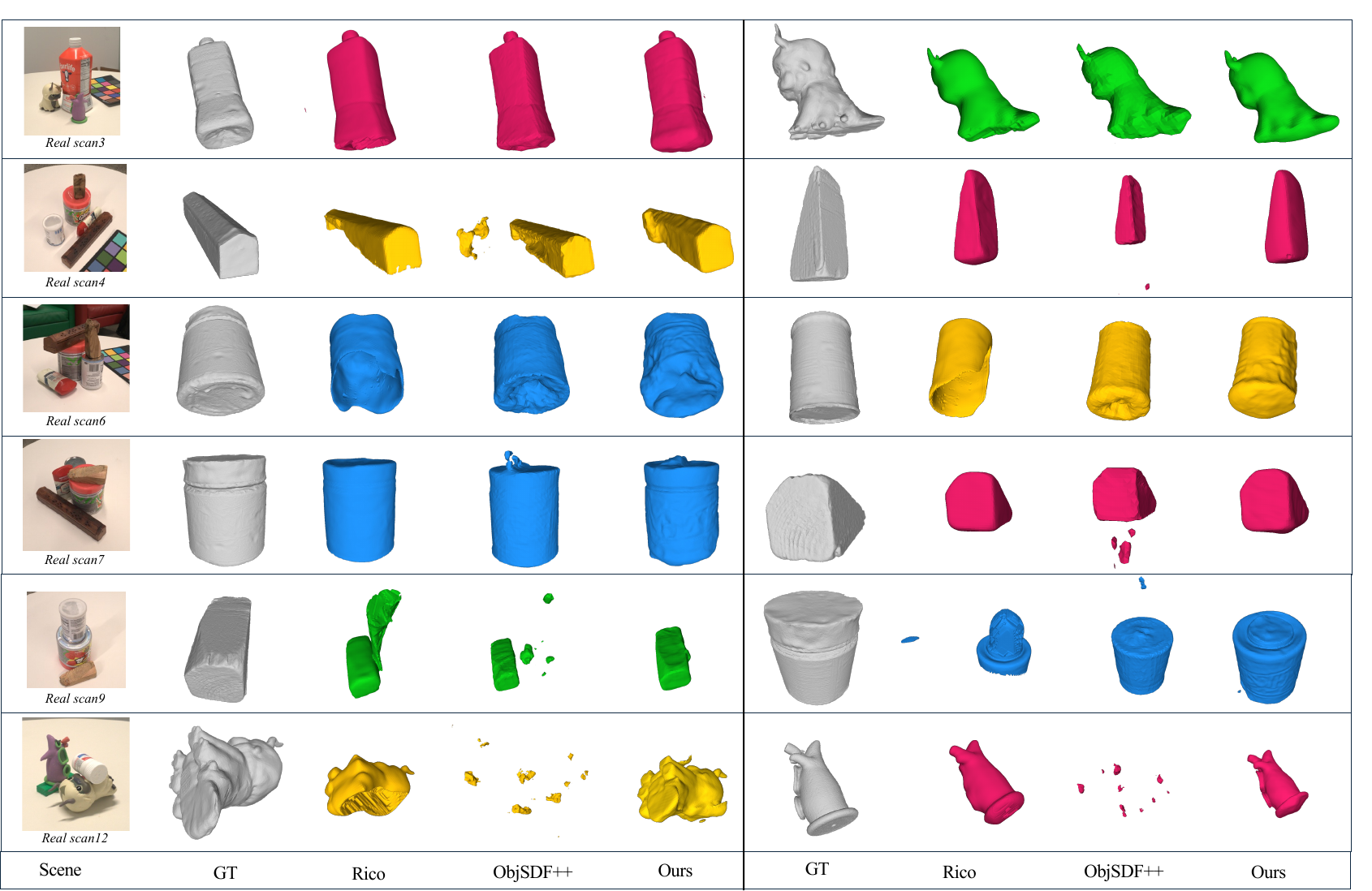}

   \caption{Comparison between RICO, ObjectSDF++ and our approach. ObjectSDF++ produces fewer details and more floating artifacts}
   \label{fig:results}
   
\end{figure*}

 We provide more qualitative results on our dataset in Figure \ref{fig:results}. We can observe that our method produces much higher quality and fewer floating artifacts compared to previous methods.

\section{Implementation details}
\subsection{Coreset Algorithm}
This algorithm takes as input a set of projected 2D points and selects $n$ points that will later be used as \textit{seed points} for SAM to segment a specific object (\textit{seed points} being the set of $(x, y)$ coordinates used to prompt SAM to segment the object). The intuition behind using this algorithm is to simulate how a human would select the seed point to segment an object using SAM. It starts by clicking the centroid of the object; the next point will be far away from the centroid, and the following point will be far away from both of the previous points. As a result, these points capture the overall shape of the object.  
We chose $n=15$ as it works for most cases.

\begin{algorithm}[H]
\caption{Modified Coreset Algorithm}\label{your_algorithm}
\begin{algorithmic}[1]
    \State \textbf{Input:} projected points $S \subset \mathbb{R}^2$, coreset size $n$
    \State \textbf{Output:} $C$
    \State $C \gets \{\}$
    \State $x_{0} \gets \arg\min_{s \in S} \|s - \text{mean}(S)\|_2$
    \State $C.\text{add}(x_{0})$
    \State $S.\text{remove}(x_{0})$
    \While{$|C| < n$}
        \State $y \gets \arg\max_{s \in S}\min\{\|s-c\|_2 : c \in C\}$
        \State $C.\text{add}(y)$
        \State $S.\text{remove}(y)$
    \EndWhile
\end{algorithmic}
\end{algorithm}

\subsection{Partial depth ordering}
When there is an overlap between two segmentation outputs from SAM, the partial depth ordering is used to break the tie. Below we describe the steps:
\subsubsection*{Step 0: Initialization}
\begin{itemize}
    \item Initialize the depth as zero for each of the $K$ objects.
\end{itemize}

\subsubsection*{Step 1: Overlap Checking}
\begin{itemize}
    \item For each pair of objects $(k_h, k_i)$ where $h \neq i$: Check if the segmentation masks of object $k_h$ and object $k_i$ overlap.
\end{itemize}

\subsubsection*{Step 2: Overlap Resolution}
\begin{itemize}
    \item If there is an overlap between the segmentation masks of objects $k_h$ and $k_i$:
    \begin{enumerate}
        \item Count the number of seed points in the overlapping region for both segmentation masks.
        \item Identify the object with more seed points in the overlapping region as the "top" object and the one with fewer seed points as the "bottom" object.
        \item Increase the depth of the top object by one relative to the depth of the bottom object.
    \end{enumerate}
\end{itemize}

{
    \small
    \bibliographystyle{ieeenat_fullname}
    \bibliography{main}

\begin{thebibliography}{39}
\providecommand{\natexlab}[1]{#1}
\providecommand{\url}[1]{\texttt{#1}}
\expandafter\ifx\csname urlstyle\endcsname\relax
  \providecommand{\doi}[1]{doi: #1}\else
  \providecommand{\doi}{doi: \begingroup \urlstyle{rm}\Url}\fi

\bibitem[Ble()]{Blender}
Blender.

\bibitem[Pol()]{Polycam}
Polycam.

\bibitem[Barron et~al.(2021{\natexlab{a}})Barron, Mildenhall, Tancik, Hedman, Martin{-}Brualla, and Srinivasan]{mipnerf}
Jonathan~T. Barron, Ben Mildenhall, Matthew Tancik, Peter Hedman, Ricardo Martin{-}Brualla, and Pratul~P. Srinivasan.
\newblock Mip-nerf: {A} multiscale representation for anti-aliasing neural radiance fields.
\newblock \emph{CoRR}, abs/2103.13415, 2021{\natexlab{a}}.

\bibitem[Barron et~al.(2021{\natexlab{b}})Barron, Mildenhall, Verbin, Srinivasan, and Hedman]{mipnerf360}
Jonathan~T. Barron, Ben Mildenhall, Dor Verbin, Pratul~P. Srinivasan, and Peter Hedman.
\newblock Mip-nerf 360: Unbounded anti-aliased neural radiance fields.
\newblock \emph{CoRR}, abs/2111.12077, 2021{\natexlab{b}}.

\bibitem[Barron et~al.(2023)Barron, Mildenhall, Verbin, Srinivasan, and Hedman]{zipnerf}
Jonathan~T Barron, Ben Mildenhall, Dor Verbin, Pratul~P Srinivasan, and Peter Hedman.
\newblock Zip-nerf: Anti-aliased grid-based neural radiance fields.
\newblock \emph{arXiv preprint arXiv:2304.06706}, 2023.

\bibitem[Breckon and Fisher(2005)]{Amodal}
Toby~P. Breckon and Robert~B. Fisher.
\newblock Amodal volume completion: 3d visual completion.
\newblock \emph{Computer Vision and Image Understanding}, 99\penalty0 (3):\penalty0 499--526, 2005.

\bibitem[Cen et~al.(2023)Cen, Zhou, Fang, Shen, Xie, Zhang, and Tian]{sa3d}
Jiazhong Cen, Zanwei Zhou, Jiemin Fang, Wei Shen, Lingxi Xie, Xiaopeng Zhang, and Qi Tian.
\newblock Segment anything in 3d with nerfs.
\newblock \emph{arXiv preprint arXiv:2304.12308}, 2023.

\bibitem[Kerr et~al.(2023)Kerr, Kim, Goldberg, Kanazawa, and Tancik]{lerf}
Justin Kerr, Chung~Min Kim, Ken Goldberg, Angjoo Kanazawa, and Matthew Tancik.
\newblock Lerf: Language embedded radiance fields, 2023.

\bibitem[Kirillov et~al.(2023{\natexlab{a}})Kirillov, Mintun, Ravi, Mao, Rolland, Gustafson, Xiao, Whitehead, Berg, Lo, Dollar, and Girshick]{kirillov2023segment}
Alexander Kirillov, Eric Mintun, Nikhila Ravi, Hanzi Mao, Chloe Rolland, Laura Gustafson, Tete Xiao, Spencer Whitehead, Alexander~C. Berg, Wan-Yen Lo, Piotr Dollar, and Ross Girshick.
\newblock Segment anything.
\newblock In \emph{ICCV}, 2023{\natexlab{a}}.

\bibitem[Kirillov et~al.(2023{\natexlab{b}})Kirillov, Mintun, Ravi, Mao, Rolland, Gustafson, Xiao, Whitehead, Berg, Lo, Dollár, and Girshick]{sam}
Alexander Kirillov, Eric Mintun, Nikhila Ravi, Hanzi Mao, Chloe Rolland, Laura Gustafson, Tete Xiao, Spencer Whitehead, Alexander~C. Berg, Wan-Yen Lo, Piotr Dollár, and Ross Girshick.
\newblock Segment anything, 2023{\natexlab{b}}.

\bibitem[Kobayashi et~al.(2022)Kobayashi, Matsumoto, and Sitzmann]{kobayashi2022decomposing}
Sosuke Kobayashi, Eiichi Matsumoto, and Vincent Sitzmann.
\newblock Decomposing nerf for editing via feature field distillation, 2022.

\bibitem[Li et~al.(2020)Li, Niklaus, Snavely, and Wang]{nsff}
Zhengqi Li, Simon Niklaus, Noah Snavely, and Oliver Wang.
\newblock Neural scene flow fields for space-time view synthesis of dynamic scenes.
\newblock \emph{CoRR}, abs/2011.13084, 2020.

\bibitem[Li et~al.(2023{\natexlab{a}})Li, Lyu, Ding, Wang, Liao, and Liu]{li2023rico}
Zizhang Li, Xiaoyang Lyu, Yuanyuan Ding, Mengmeng Wang, Yiyi Liao, and Yong Liu.
\newblock Rico: Regularizing the unobservable for indoor compositional reconstruction, 2023{\natexlab{a}}.

\bibitem[Li et~al.(2023{\natexlab{b}})Li, Wang, Cole, Tucker, and Snavely]{dynibar}
Zhengqi Li, Qianqian Wang, Forrester Cole, Richard Tucker, and Noah Snavely.
\newblock Dynibar: Neural dynamic image-based rendering.
\newblock In \emph{Proceedings of the IEEE/CVF Conference on Computer Vision and Pattern Recognition}, pages 4273--4284, 2023{\natexlab{b}}.

\bibitem[Mescheder et~al.(2018)Mescheder, Oechsle, Niemeyer, Nowozin, and Geiger]{OccupancyNet}
Lars~M. Mescheder, Michael Oechsle, Michael Niemeyer, Sebastian Nowozin, and Andreas Geiger.
\newblock Occupancy networks: Learning 3d reconstruction in function space.
\newblock \emph{CoRR}, abs/1812.03828, 2018.

\bibitem[Mildenhall et~al.(2020)Mildenhall, Srinivasan, Tancik, Barron, Ramamoorthi, and Ng]{NeRF}
Ben Mildenhall, Pratul~P. Srinivasan, Matthew Tancik, Jonathan~T. Barron, Ravi Ramamoorthi, and Ren Ng.
\newblock Nerf: Representing scenes as neural radiance fields for view synthesis.
\newblock \emph{CoRR}, abs/2003.08934, 2020.

\bibitem[Monnier et~al.(2023)Monnier, Austin, Kanazawa, Efros, and Aubry]{dbw}
Tom Monnier, Jake Austin, Angjoo Kanazawa, Alexei~A. Efros, and Mathieu Aubry.
\newblock Differentiable blocks world: Qualitative 3d decomposition by rendering primitives, 2023.

\bibitem[M{\"{u}}ller et~al.(2022)M{\"{u}}ller, Evans, Schied, and Keller]{iNGP}
Thomas M{\"{u}}ller, Alex Evans, Christoph Schied, and Alexander Keller.
\newblock Instant neural graphics primitives with a multiresolution hash encoding.
\newblock \emph{CoRR}, abs/2201.05989, 2022.

\bibitem[Niemeyer and Geiger(2021)]{giraffe}
Michael Niemeyer and Andreas Geiger.
\newblock Giraffe: Representing scenes as compositional generative neural feature fields.
\newblock In \emph{CVPR}, 2021.

\bibitem[Niemeyer et~al.(2019)Niemeyer, Mescheder, Oechsle, and Geiger]{DVR}
Michael Niemeyer, Lars~M. Mescheder, Michael Oechsle, and Andreas Geiger.
\newblock Differentiable volumetric rendering: Learning implicit 3d representations without 3d supervision.
\newblock \emph{CoRR}, abs/1912.07372, 2019.

\bibitem[Park et~al.(2019)Park, Florence, Straub, Newcombe, and Lovegrove]{DeepSDF}
Jeong~Joon Park, Peter~R. Florence, Julian Straub, Richard~A. Newcombe, and Steven Lovegrove.
\newblock Deepsdf: Learning continuous signed distance functions for shape representation.
\newblock \emph{CoRR}, abs/1901.05103, 2019.

\bibitem[Park et~al.(2020)Park, Sinha, Barron, Bouaziz, Goldman, Seitz, and Martin{-}Brualla]{nerfies}
Keunhong Park, Utkarsh Sinha, Jonathan~T. Barron, Sofien Bouaziz, Dan~B. Goldman, Steven~M. Seitz, and Ricardo Martin{-}Brualla.
\newblock Deformable neural radiance fields.
\newblock \emph{CoRR}, abs/2011.12948, 2020.

\bibitem[Park et~al.(2021)Park, Sinha, Hedman, Barron, Bouaziz, Goldman, Martin{-}Brualla, and Seitz]{hypernerf}
Keunhong Park, Utkarsh Sinha, Peter Hedman, Jonathan~T. Barron, Sofien Bouaziz, Dan~B. Goldman, Ricardo Martin{-}Brualla, and Steven~M. Seitz.
\newblock Hypernerf: {A} higher-dimensional representation for topologically varying neural radiance fields.
\newblock \emph{CoRR}, abs/2106.13228, 2021.

\bibitem[Reiser et~al.(2021)Reiser, Peng, Liao, and Geiger]{KiloNeRF}
Christian Reiser, Songyou Peng, Yiyi Liao, and Andreas Geiger.
\newblock Kilonerf: Speeding up neural radiance fields with thousands of tiny mlps.
\newblock \emph{CoRR}, abs/2103.13744, 2021.

\bibitem[Ren et~al.(2022)Ren, Agarwala$^\dagger$, Russell$^\dagger$, Schwing$^\dagger$, and Wang$^\dagger$]{ren2022nvos}
Zhongzheng Ren, Aseem Agarwala$^\dagger$, Bryan Russell$^\dagger$, Alexander~G. Schwing$^\dagger$, and Oliver Wang$^\dagger$.
\newblock Neural volumetric object selection.
\newblock In \emph{IEEE/CVF Conference on Computer Vision and Pattern Recognition (CVPR)}, 2022.
\newblock ($^\dagger$ alphabetic ordering).

\bibitem[Rosu and Behnke(2023)]{rosu2023permutosdf}
Radu~Alexandru Rosu and Sven Behnke.
\newblock Permutosdf: Fast multi-view reconstruction with implicit surfaces using permutohedral lattices, 2023.

\bibitem[Schonberger and Frahm(2016)]{Schonberger_2016_CVPR}
Johannes~L. Schonberger and Jan-Michael Frahm.
\newblock Structure-from-motion revisited.
\newblock In \emph{Proceedings of the IEEE Conference on Computer Vision and Pattern Recognition (CVPR)}, 2016.

\bibitem[Takikawa et~al.(2021)Takikawa, Litalien, Yin, Kreis, Loop, Nowrouzezahrai, Jacobson, McGuire, and Fidler]{NGLOD}
Towaki Takikawa, Joey Litalien, Kangxue Yin, Karsten Kreis, Charles~T. Loop, Derek Nowrouzezahrai, Alec Jacobson, Morgan McGuire, and Sanja Fidler.
\newblock Neural geometric level of detail: Real-time rendering with implicit 3d shapes.
\newblock \emph{CoRR}, abs/2101.10994, 2021.

\bibitem[Wang et~al.(2021)Wang, Liu, Liu, Theobalt, Komura, and Wang]{neus}
Peng Wang, Lingjie Liu, Yuan Liu, Christian Theobalt, Taku Komura, and Wenping Wang.
\newblock Neus: Learning neural implicit surfaces by volume rendering for multi-view reconstruction.
\newblock \emph{CoRR}, abs/2106.10689, 2021.

\bibitem[Wu et~al.(2022)Wu, Liu, Chen, Li, Zheng, Cai, and Zheng]{objsdf}
Qianyi Wu, Xian Liu, Yuedong Chen, Kejie Li, Chuanxia Zheng, Jianfei Cai, and Jianmin Zheng.
\newblock Object-compositional neural implicit surfaces, 2022.

\bibitem[Wu et~al.(2023)Wu, Wang, Li, Zheng, and Cai]{objsdf++}
Qianyi Wu, Kaisiyuan Wang, Kejie Li, Jianmin Zheng, and Jianfei Cai.
\newblock Objectsdf++: Improved object-compositional neural implicit surfaces, 2023.

\bibitem[Xu et~al.(2022)Xu, Chai, Shi, Peng, Skorokhodov, Siarohin, Yang, Shen, Lee, Zhou, and Tulyakov]{xu2022discoscene}
Yinghao Xu, Menglei Chai, Zifan Shi, Sida Peng, Ivan Skorokhodov, Aliaksandr Siarohin, Ceyuan Yang, Yujun Shen, Hsin-Ying Lee, Bolei Zhou, and Sergey Tulyakov.
\newblock Discoscene: Spatially disentangled generative radiance fields for controllable 3d-aware scene synthesis, 2022.

\bibitem[Yang et~al.(2021)Yang, Zhang, Xu, Li, Zhou, Bao, Zhang, and Cui]{objcompnerf}
Bangbang Yang, Yinda Zhang, Yinghao Xu, Yijin Li, Han Zhou, Hujun Bao, Guofeng Zhang, and Zhaopeng Cui.
\newblock Learning object-compositional neural radiance field for editable scene rendering.
\newblock \emph{CoRR}, abs/2109.01847, 2021.

\bibitem[Yariv et~al.(2020)Yariv, Atzmon, and Lipman]{IDR}
Lior Yariv, Matan Atzmon, and Yaron Lipman.
\newblock Universal differentiable renderer for implicit neural representations.
\newblock \emph{CoRR}, abs/2003.09852, 2020.

\bibitem[Yariv et~al.(2021)Yariv, Gu, Kasten, and Lipman]{volsdf}
Lior Yariv, Jiatao Gu, Yoni Kasten, and Yaron Lipman.
\newblock Volume rendering of neural implicit surfaces.
\newblock \emph{CoRR}, abs/2106.12052, 2021.

\bibitem[Yu et~al.(2021)Yu, Li, Tancik, Li, Ng, and Kanazawa]{PlenOctrees}
Alex Yu, Ruilong Li, Matthew Tancik, Hao Li, Ren Ng, and Angjoo Kanazawa.
\newblock Plenoctrees for real-time rendering of neural radiance fields.
\newblock \emph{CoRR}, abs/2103.14024, 2021.

\bibitem[Yu et~al.(2022{\natexlab{a}})Yu, Guibas, and Wu]{yu2022unsupervised}
Hong-Xing Yu, Leonidas~J. Guibas, and Jiajun Wu.
\newblock Unsupervised discovery of object radiance fields.
\newblock In \emph{ICLR}, 2022{\natexlab{a}}.

\bibitem[Yu et~al.(2022{\natexlab{b}})Yu, Peng, Niemeyer, Sattler, and Geiger]{monosdf}
Zehao Yu, Songyou Peng, Michael Niemeyer, Torsten Sattler, and Andreas Geiger.
\newblock Monosdf: Exploring monocular geometric cues for neural implicit surface reconstruction, 2022{\natexlab{b}}.

\bibitem[Zhang et~al.(2020)Zhang, Riegler, Snavely, and Koltun]{nerf++}
Kai Zhang, Gernot Riegler, Noah Snavely, and Vladlen Koltun.
\newblock Nerf++: Analyzing and improving neural radiance fields.
\newblock \emph{arXiv preprint arXiv:2010.07492}, 2020.

\end{thebibliography}
}

\end{document}